\documentclass[10pt,twocolumn,letterpaper]{article}

\usepackage{iccv}
\usepackage{times}
\usepackage{epsfig}
\usepackage{graphicx}
\usepackage{amsmath}
\usepackage{amssymb}

\usepackage{multirow}
\usepackage{subfig}
\usepackage{makecell}
\usepackage{booktabs}
\usepackage{soul}
\soulregister\cite7
\soulregister\ref7

\usepackage[pagebackref=true,breaklinks=true,letterpaper=true,colorlinks,bookmarks=false]{hyperref}

\iccvfinalcopy % *** Uncomment this line for the final submission

\def\iccvPaperID{2163} % *** Enter the ICCV Paper ID here

\ificcvfinal\fi

\begin{document}

%%%%%%%%% TITLE
\title{HSR-Diff: Hyperspectral Image Super-Resolution via Conditional Diffusion Models}

% \author{Chanyue Wu\\
% Northwestern Polytechnical University \\
%  P.R.China.\\
% {\tt\small chanyuewu@mail.nwpu.edu.cn}

% % \and
% % Dong Wang\\
% % Yan’an University \\
% % % 580 Shendi Road, Yan'an Shaanxi,761000, P.R.China.\\
% % Yan'an Shaanxi, China.\\
% % {\tt\small dongwang@mail.nwpu.edu.cn}

% % \and
% % Hanyu Mao \\
% % Northwestern Polytechnical University \\
% % 1 Dongxiang Road, Chang'an District, Xi'an Shaanxi,710129, P.R.China.\\
% % {\tt\small maomhy@mail.nwpu.edu.cn}

% % \and
% % Ying Li\\
% % Northwestern Polytechnical University \\
% % 1 Dongxiang Road, Chang'an District, Xi'an Shaanxi,710129, P.R.China.\\
% % {\tt\small lybyp@nwpu.edu.cn}

% }

\author{
Chanyue Wu\\
Northwestern Polytechnical University \\
Xi'an, Shaanxi, China\\
{\tt\small chanyuewu@mail.nwpu.edu.cn}
\and
Dong Wang\\
Yan’an University \\
Yan'an, Shaanxi, China\\
{\tt\small dongwang@mail.nwpu.edu.cn}
\and
Hanyu Mao \\
Northwestern Polytechnical University \\
Xi'an, Shaanxi, China\\
{\tt\small maomhy@mail.nwpu.edu.cn}
\and
Ying Li\\
Northwestern Polytechnical University \\
Xi'an, Shaanxi, China\\
{\tt\small lybyp@nwpu.edu.cn}
}

\maketitle
\ificcvfinal\thispagestyle{empty}\fi

%%%%%%%%% ABSTRACT
\begin{abstract}
    % Due to the inevitable trade-off between spatial and spectral resolution, hyperspectral imaging systems can directly collect either high-resolution (HR) multispectral image (MSI) or low-resolution (LR) hyperspectral image (HSI).
    % Fusion-based Hyperspectral image (HSI) super-resolution (SR) aims to generate HR HSI by merging HR-MSI and LR-HSI.
    % To the best of our knowledge, this is the first study with diffusion probabilistic models in the field of HSI-SR.
    % HSR-Diff initializes the HR-HSI with pure Gaussian noise and iteratively refines it with the condition of the LR-HSI and the HR-MSI.
% \hl    {At each step, the noise is removed with a Conditional Denoising Transformer (CDFormer) which utilizes the hierarchical representations of HR-MSI and LR-HSI rather than the original images.}
    % Extensive experiments are conducted on four public datasets to validate the superiority of the proposed method when compared with other state-of-the-art ones.
    % Systematic experiments have been conducted on four public datasets, demonstrating that the proposed HSR-Diff outperforms state-of-the-art methods.
    Despite the proven significance of hyperspectral images (HSIs) in performing various computer vision tasks, its potential is adversely affected by the low-resolution (LR) property in the spatial domain, resulting from multiple physical factors.
    Inspired by recent advancements in deep generative models, we propose an HSI Super-resolution (SR) approach with Conditional Diffusion Models (HSR-Diff) that merges a high-resolution (HR) multispectral image (MSI) with the corresponding LR-HSI.
    HSR-Diff generates an HR-HSI via repeated refinement, in which the HR-HSI is initialized with pure Gaussian noise and iteratively refined.
    At each iteration, the noise is removed with a Conditional Denoising Transformer (CDFormer) that is trained on denoising at different noise levels, conditioned on the hierarchical feature maps of HR-MSI and LR-HSI.
    In addition, a progressive learning strategy is employed to exploit the global information of full-resolution images.
    Systematic experiments have been conducted on four public datasets, demonstrating that HSR-Diff outperforms state-of-the-art methods.
\end{abstract}

\vspace{-0.811mm}
\section{Introduction}
\begin{figure}
    \centering
    \includegraphics[width=0.8\linewidth]{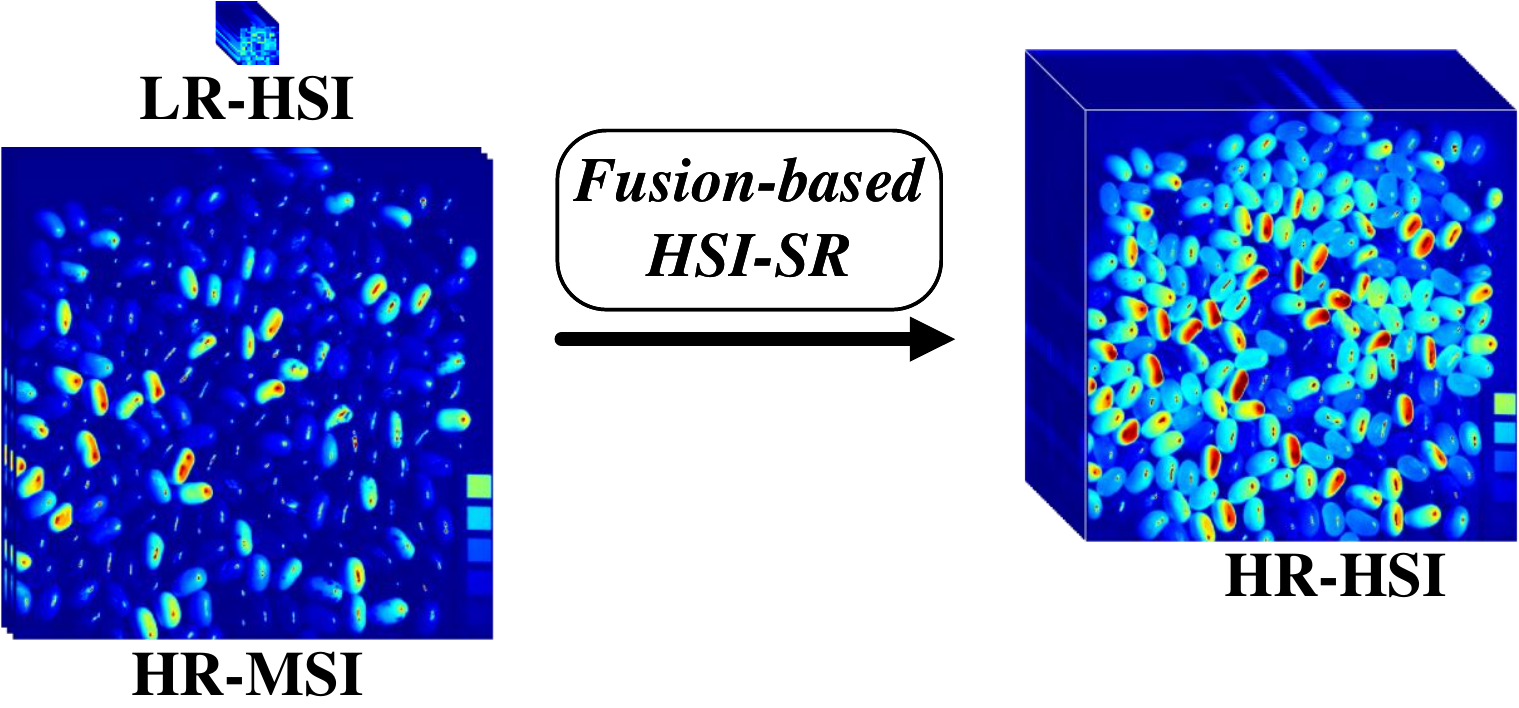}
    \caption{Illustration of HSI super-resolution.}
    \label{Fig:Fusion-based-HSI-SR}
    \vspace{-3mm}
\end{figure}

% The main objective of hyperspectral imaging is to provide accurate images of a targeted scene.
% The collected hyperspectral images (HSI) contain rich spectral information delivered by dozens or hundreds of spectral bands.
% Benefiting from the rich spectral information conveyed by the densely sampled spectra, HSIs provide more faithful knowledge of targeted scenes than conventional imaging modalities.
% As a result, HSIs play an irreplaceable role in various tasks, e.g., HSI classification \cite{jiang2021hyperspectral}, \hl{target detection }\cite{jiang2021e2e}, and so on.
% Besides spectral resolution, spatial resolution is another essential property of HSIs that significantly impacts the performance of downstream applications.
% High spatial resolution (HR) can be obtained by improving the hyperspectral instruments to resolve many more details in the space dsomain.
% However, it turns out to be a challenging task to directly improve the spatial resolution due to the stringent constraint of the signal-to-noise ratio of captured HSIs.
% In this regard, a simple solution of using high resolution sensors is not viable as it further reduces the density of the photons reaching the sensors, which is already limited by the high spectral resolution of the instruments.
% A practical way around the problem consists of fusing the low spatial resolution (LR) HSI with an HR multispectral image (MSI), the technique of which is referred to as fusion-based HSI super-resolution (SR).
Hyperspectral images (HSI) contain dozens or hundreds of spectral bands, enabling them to provide more faithful knowledge of targeted scenes than conventional imaging modalities.
As such, HSIs play an irreplaceable role in various computer vision tasks, including classification \cite{xue2022grafting,zhang20213}, segmentation \cite{dao2021improving}, and tracking \cite{xiong2020material}.
% Although HSIs contain rich spectral information, contemporary hyperspectral imaging sensors lack high-resolution (HR) in the spatial domain due to the stringent constraint of the signal-to-noise ratio of the final image product.
Although HSIs contain rich spectral information, contemporary hyperspectral imaging sensors lack high-resolution (HR) in the spatial domain, due to the stringent constraint of typically low signal-to-noise ratios.
Their widespread use is significantly hindered by this fact.
Restricted by hardware limitations, a practical way to work around this problem is to fuse the low-resolution (LR) HSI with an HR multispectral image (MSI).
% This requires the implementation of so-called HSI super-resolution (SR), or HSI-SR for short, as shown in Figure \ref{Fig:Fusion-based-HSI-SR}. % or hyperspectral and multispectral image fusion.
This requires the implementation of so-called HSI super-resolution (SR), as shown in Figure \ref{Fig:Fusion-based-HSI-SR}. % or hyperspectral and multispectral image fusion.

% Inspired by the similarity between pansharpening and HSI-SR, 
% Such approaches are easy to implement 
Over the past few decades, a significant amount of research efforts have been devoted to developing HSI-SR approaches, which can be roughly classified into five categories \cite{zhang2021survey}: Extensions of pansharpening \cite{chen2014fusion}, Bayesian inference-based \cite{akhtar2015bayesian,wei2015hyperspectral}, matrix factorization-based \cite{borsoi2019super}, tensor-based \cite{li2018fusing}, and deep learning (DL)-based.
Whilst pansharpening methods \cite{wang2022convolutional} have been extended to the field of HSI-SR, such approaches are prone to spectral distortion.
Bayesian inference-based approaches rely on the assumption of prior knowledge, thereby having a weak flexibility in dealing with different HSI structures.
Matrix factorization-based techniques reshape the 3D HSIs and MSIs into matrices, thus facing the challenge of learning the required relationship between space and spectrum.
Although several tensor-based methods have been proposed that can maintain the 3D structure of input images, they consume much more memory and computational power.
Furthermore, these traditional approaches work via relying heavily on hand-crafted priors.

% Since representation can maintain the 3D structure of input images, several tensor-based approaches have been proposed.
% However, compared with Matrix factorization-based approaches, tensor-based methods require more computation resources.
% Unlike the other methods relying on hand-crafted features, DL-based approaches learn prior knowledge automatically from given data.
% As one kind of deep learning technique, deep generative models have seen success in learning complex empirical distributions of images.
% \hl{Among them, deep generative models, such as autoregressive models (AR), generative adversarial networks (GAN), variational autoencoders (VAE), normalizing flows (NF), and diffusion probabilistic models (``diffusion model" for brevity), have seen success in learning complex empirical distributions of HSIs and have shown convincing image generation results\cite{shi2022latent}.}
% However, GANs require carefully designed regularization and optimization tricks to tame optimization instability and mode collapse.
% Autoregressive models are computationally expensive for high-resolution image generation.
% NFs and VAEs often yield sub-optimal sample quality.
% In addition, CNN-based approaches generally show limitations for the modeling of long-range dependencies.
% Furthermore, CNNs are usually trained on image patches, which results in suboptimal performance on full-resolution images at test time.
Recently, DL-based methods, especially convolutional neural network (CNN)-based approaches, have flooded over into the HSI-SR research community \cite{dian2018deep,xie2020mhf,fu2019hyperspectral,qu2018unsupervised,zheng2020coupled,zhang2020unsupervised,li2017spectral}.
Rather than resorting to hand-crafted features, DL-based techniques learn prior knowledge automatically from given data.
Particularly, Dong et al. proposed the first DL-based method for image SR, with the end-to-end mapping between LR images and HR images learned using a CNN \cite{dong2015image}.
Subsequently, generative adversarial networks (GANs) were introduced to the field of image SR in an effort to produce high-frequency details \cite{ledig2017photo,goodfellow2020generative}.
After that, various GAN-based models have been devised, showing state-of-the-art results in the HSI-SR literature \cite{shi2022latent}.
However, such work requires carefully designed regularization and optimization tricks to tame optimization instability and avoid mode collapse.

% We adapt DDPMs to the HSI-SR task by proposing a simple effective modification to 
% Denoising Diffusion Probabilistic Models (DDPMs)
% Inspired by the recent developments in deep generative models, we propose a Diffusion probabilistic Model for HSI-SR.
% To be specific, the HR-HSI is initialized with pure Gaussian noise and iteratively refined.
Inspired by the recent developments in deep generative models, in this paper, we propose an innovative approach that we refer to as HSR-Diff (HSI-SR with conditional diffusion models).
It works by learning to transform the original standard normal distribution into the data distribution of HR-HSI through a sequence of refinements.
In contrast to GAN-based methods which require inner-loop maximization, HSR-Diff minimizes a well-defined loss function.
Although conditional diffusion models are straightforward to define and efficient to train, there has been no demonstration that they are capable of merging LR-HSI and HR-MSI to the best of our knowledge.
We show that conditional diffusion models are capable of generating high-quality HR-HSIs, which may best the
 state-of-the-art results.
A key factor of HSR-Diff is its inherent denoising ability thanks to use of deep neural networks.
In spite of the effectiveness of CNNs for denoising, they have shown limitation in modelling long-range dependencies.
To address the locality problem of convolution operations, a Conditional Denoising Transformer (CDFormer) is herein designed and trained with a denoising objective to remove various levels of noise iteratively.
In addition, a progressive learning strategy is utilized to help the CDFormer learn the global statistics of full-resolution HSIs.
The main contributions of this work are summarized as follows:

\begin{itemize}
    
    \item %We propose an HSI-SR approach based on Diffusion Models.
    We propose the novel application of conditional diffusion models in the field of HSI-SR that works by progressively destroying HR-HSI through injecting noise and subsequently learning to reverse this process, in order to perform HR-HSI.
    % To the best of our knowledge, this is the first work to exploit diffusion models in the field of HSI-SR.
    % With the aid of repeated refinement, HSR-Diff restore 
    % HSR-Diff iniialize the HR HSI with pure Gaussian noise and iteratively refine it conditioned on the LR-HSI and the HR-MSI.
    % HSR-Diff progressively destroys HR-HSI by injecting noise, then learns to reverse this process for HR-HSI generation.
    
    \item 
    % We design a CDFormer that refines a noisy HR-HSI conditioned on the deep feature maps of HR-MSI and LR-HSI.
    % The CDFormer consists of an SR stream and a denoising stream, where the former extracts hierarchical features of HR-MSI and LR-HSI, while the latter generates a high-quality HR-HSI from a noisy one, conditioned on the noise level and image representations.
    We introduce a CDFormer that refines a noisy HR-HSI conditioned on the deep feature maps of HR-MSI and LR-HSI, capable of modelling global connectivity with a self-attention mechanism.
    % CDFormer is capable of modelling global connectivity with the self-attention mechanism.
    
    \item We employ a progressive learning strategy to exploit the global information of full-resolution HSIs, with CDFormer being trained on small image patches in the early epochs with high efficiency and on the global images in the later epochs to acquire global information.
    
    \item We present experimental investigations on four public datasets, with quantitative and qualitative results illustrating the superior performance of our approach as compared with state-of-the-art methods.

\end{itemize}

\vspace{-0.811mm}
\section{Related Work}

\vspace{-0.622mm}
\subsection{Deep Generative Models}
\vspace{-0.222mm}

% autoregressive models (AR), generative adversarial networks (GAN), variational autoencoders (VAE), normalizing flows (NF), and diffusion probabilistic models
Typical deep generative models include autoregressive models (AR), normalizing flows (NF), variational autoencoders (VAE), GANs, and diffusion models.
ARs learn data distribution via log-likelihood.
Unfortunately, the low efficiency of sequential pixel generation limits their application to high-resolution images \cite{van2016pixel,salimans2017pixelcnn++}.
NFs have the advantage of running at a high sampling speed, but their expressive ability is restricted by the need for invertible parameterized transformations with a tractable Jacobian determinant \cite{rezende2015variational,dinh2016density,kingma2018glow}.
VAEs feature fast sampling while underperforming in comparison to GANs and ARs, in terms of image quality \cite{kingma2013auto,rezende2014stochastic,vahdat2020nvae}.
GANs are popular for class conditional image generation, and super-resolution \cite{goodfellow2020generative}.
However, the inner-outer loop optimization often requires tricks to stabilize training \cite{arjovsky2017wasserstein,gulrajani2017improved}, and conditional tasks like super-resolution usually demand an auxiliary consistency-based loss to avoid mode collapse \cite{ledig2017photo}.

The development of diffusion models has seen a dramatically accelerating pace over the past three years. 
% \hl{In this paper, we extend to the field of HSI-SR the diffusion model, which is a new kind of deep generative model.}
Whilst diffusion models have shown great potential for a variety of computer vision applications, 
none of them have yet been devoted to the problem of HSI-SR to the best of our knowledge. 
% In this paper, we extend to the field of HSI-SR the diffusion model.
In this paper, we extend the utility of diffusion models to the field of HSI-SR.

\vspace{-0.622mm}
\subsection{Deep Learning-Based HSI-SR}
\vspace{-0.222mm}
% In this section, we introduce some DL-based HSI-SR approaches to which category our proposed method belongs.
% In this section, we introduce some DL-based HSI-SR approaches.
In recent years, data-driven CNN architectures have been shown to outperform traditional approaches for use in the HSI-SR literature.
These methods formulate the underlying fusion problem as a highly nonlinear mapping that takes HR-MSIs and LR-HSIs as input to generate an optimal HR-HSI.
For example, CMHF-net \cite{xie2020mhf} is an interpretable CNN, the design of which exploits the deep unfolding technique.
Zhang et al. \cite{zhang2020unsupervised} proposed to reconstruct HR-HSIs with a two-stage network, while Zhang et al. \cite{zhang2020ssr} designed an interpretable spatial-spectral reconstruction network (SSR-NET) based on CNN.
Aiming at problems of inflexible structure and information distortion, Jin et al. embedded Bilateral Activation Mechanism into ResNet, resulting in the effective model of BRResNet \cite{jin2021bam}.
Thanks to the inductive bias of CNN, such as locality and weight sharing, these methods can provide good generalization performance and achieve impressive results.

Nevertheless, CNNs have limitations in capturing long-range dependencies and self-similarity priors.
To overcome such shortcomings we employ Transformer to learn global statistics of full-resolution images in this work.

\begin{figure}
    \centering
    \includegraphics[width=\linewidth]{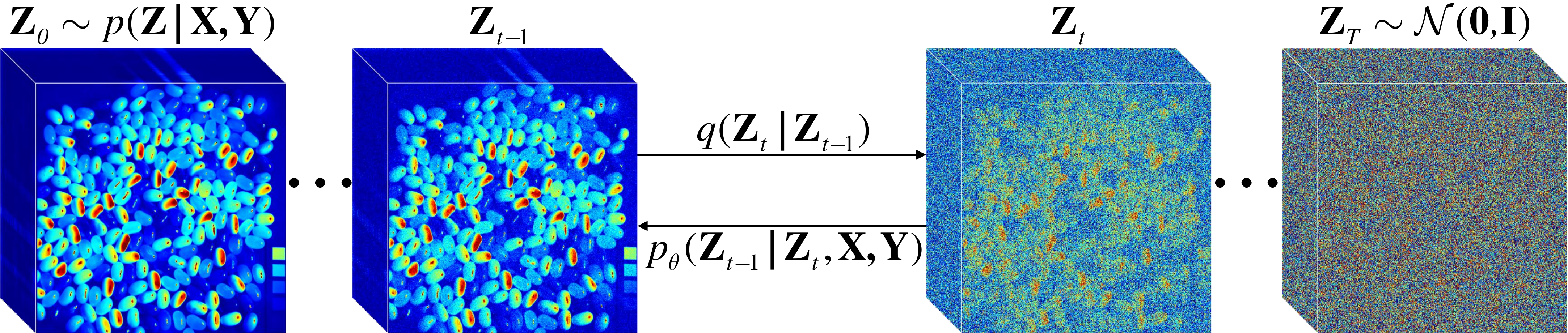}
    \caption{Forward and reverse processes of HSR-Diff, with forward process $q$ generating an HSI sequence (left to right) by gradually adding Gaussian noise, and reverse process $p$ iteratively refining HR-HSI (right to left).}
    \label{Fig:Forward-diffusion-and-reverse-process}
    \vspace{-2.1mm}
\end{figure}

\vspace{-0.811mm}
\section{Proposed Methodology}

% In this section, the detailed process of our proposed *** will be elaborated step by setp.
% In summary, the *** consists of *** parts, i.e., 

% To be specific, HSR-Diff makes use of two Markov chains: a forward chain that perturbs data to noise, and a reverse chain that converts noise back to data.
% The former is typically hand-designed with the goal to transform any data distribution into a simple prior distribution (e.g., standard Gaussian), while the latter Markov chain reverses the former by learning transition kernels parameterized by deep neural networks.
% New data points are subsequently generated by first sampling a random vector from the prior distribution, followed by ancestral sampling through the reverse Markov chain [125].

\vspace{-0.622mm}
\subsection{Problem Formulation}
\vspace{-0.222mm}

% The observation models for the HR-MSI and the LR-HSI are the inverse of the fusion process, which can be mathematically formulated as
Without losing generality, the observation models for the HR-MSI and LR-HSI of interest can be mathematically formulated as
% \begin{equation}
%     \begin{aligned}
%         \mathbf{X}&=\mathbf{R}\mathbf{Z},\\
%         \mathbf{Y}&=\mathbf{Z}\mathbf{D},
%     \end{aligned}
% \end{equation}
\begin{equation}
    \mathbf{X}=\mathbf{R}\mathbf{Z},\mathbf{Y}=\mathbf{Z}\mathbf{D},
\end{equation}
where $\mathbf{X}\in\mathbb{R}^{b \times HW}$ denotes the HR-MSI which consists of $b$ spectral bands with a spatial resolution of $HW$ in the spatial domain; $\mathbf{R}\in\mathbb{R}^{b\times B}$ represents the spectral response function of HR-MSI; $\mathbf{Y}\in \mathbb{R}^{B\times hw}$ denotes the LR-HSI; and $\mathbf{Z}\in\mathbb{R}^{B \times HW}$ is the latent HR-HSI.
In the above, $b$ and $B$ are the numbers of bands, with $h$ and $H$ being the band height, and $w$ and $W$ the width, where $b\ll B$, $h \ll H$, and $w \ll W$.
$\mathbf{D}\in\mathbb{R}^{HW \times hw}$ is the spatial response of the LR-HSI, which can be modelled with blurring and down-sampling operations.
The HSI-SR can be interpreted as an inverse problem for merging a practically collected $\mathbf{X}$ and an observed $\mathbf{Y}$ to produce a latent $\mathbf{Z}$.
In ths paper, the ideal $\mathbf{Z}$ is restored with HSR-Diff conditioned on spatio-spectral information of $\mathbf{X}$ and $\mathbf{Y}$, the details of which are described below.

\vspace{-0.622mm}
\subsection{HSI-SR with Conditional Diffusion Models}
\vspace{-0.222mm}

Given a dataset $\mathcal{D}_{train}=\{\mathbf{X}^i,\mathbf{Y}^i,\mathbf{Z}^i\}_{i=1}^N$ satisfying a certain joint probability distribution $p\left(\mathbf{X},\mathbf{Y},\mathbf{Z}\right)$, many pairs of $\left(\mathbf{X},\mathbf{Y}\right)$ may be consistent with the same $\mathbf{Z}$.
Thus, the HR-HSI $\mathbf{Z}$ can be obtained with iterative refinements that provide an approximate to $p\left(\mathbf{Z}|\mathbf{X},\mathbf{Y}\right)$.
In this work, we implement the process of iterative refinements with HSR-Diff, where the optimized HR-HSI is presumed to be produced in $T$ refinement steps.
In HSR-Diff, the target HR-HSI is initialized with a pure noise $\mathbf{Z}_T\sim\mathcal{N}(\mathbf{0},\mathbf{I})$ as shown in Figure \ref{Fig:Forward-diffusion-and-reverse-process}.
The HSI is then refined iteratively according to learned conditional distributions $p_\theta\left(\mathbf{Z}_{t-1}|\mathbf{Z}_{t},\mathbf{X},\mathbf{Y}\right)$.
In so doing, the image sequence $(\mathbf{Z}_{T-1}, \mathbf{Z}_{T-2}, \ldots, \mathbf{Z}_{0})$ can be attained and ultimately $\mathbf{Z}_{0} \sim p\left(\mathbf{Z}|\mathbf{X},\mathbf{Y}\right)$.

The HSR-Diff employed makes use of two processes: a forward process that perturbs HSI to noise, and a reverse process converting noise back to HSI.
% The intermediate images, i.e., $\mathbf{Z}_{T-1}$, $\mathbf{Z}_{T-2}$, $\cdots$, and $\mathbf{Z}_{1}$, are generated according to a forward process which can be formulated as Markov chain with fixed transition probability $q\left(\mathbf{Z}_{t}|\mathbf{Z}_{t-1}\right)$.
In the forward process, the intermediate images, i.e., $\mathbf{Z}_{T-1}$, $\mathbf{Z}_{T-2}$, $\cdots$, and $\mathbf{Z}_{1}$, are generated according to a Markov chain with fixed transition probability $q\left(\mathbf{Z}_{t}|\mathbf{Z}_{t-1}\right)$.
We are interested in reversing the process via iterative refinements, in which the noise is reduced iteratively with a reverse Markov chain conditioned on $\mathbf{X}$ and $\mathbf{Y}$.
% The reverse chain is learned with the CDFormer $f_\theta$ that takes as input a noisy HR-HSI $\mathbf{Z}_{t}$, an HR-MSI $\mathbf{X}$ and an LR-HSI $\mathbf{Y}$ and estimates the noiseless HR-HSI $\mathbf{Z}_{0}$.
% In particular, the noise schedule $\alpha_{1:T}$ is a hyper-parameter of our method.
The reverse chain is learned with the CDFormer $f_\theta$.
Further details of HSR-Diff’s working are given below.

\vspace{-2.833mm}
\subsubsection{Forward Process}
\vspace{-0.5mm}

Inspired by \cite{ho2020denoising}, forward process $q$ iteratively adds Gaussian noise to $\mathbf{Z}_0$ over $T$ iterations:
\begin{equation}
    \begin{aligned}
    q\left(\mathbf{Z}_{1:T} \mid \mathbf{Z}_0\right) & =\prod_{t=1}^T q\left(\mathbf{Z}_t \mid \mathbf{Z}_{t-1}\right), \\
    q\left(\mathbf{Z}_t \mid \mathbf{Z}_{t-1}\right) & =\mathcal{N}\left(\mathbf{Z}_t ; \sqrt{\alpha_t} \mathbf{Z}_{t-1},\left(1-\alpha_t\right) \mathbf{I}\right),
    \end{aligned}
\end{equation}
where $\alpha_{1:T} \in (0,1)$ are scalar hyper-parameters.
Note that in the forward process, the distribution of $\mathbf{Z}_{t}$ given $\mathbf{Z}_{0}$ can be directly sampled in closed form.
This implies that
\begin{equation}
    q\left(\mathbf{Z}_t \mid \mathbf{Z}_0\right)=\mathcal{N}\left(\mathbf{Z}_t ; \sqrt{\gamma_t} \mathbf{Z}_0,\left(1-\gamma_t\right) \mathbf{I}\right)
\end{equation}
where $\gamma=\prod_{i=1}^t\alpha_i$.
In addition, the posterior distribution of $\mathbf{Z}_{t-1}$ given $\mathbf{Z}_{0}$ and $\mathbf{Z}_{t}$ can be derived by
\begin{equation}
    \begin{aligned}
    & q\left(\mathbf{Z}_{t-1} \mid \mathbf{Z}_0, \mathbf{Z}_t\right)=\mathcal{N}\left(\mathbf{Z}_{t-1} ; \boldsymbol{\mu}, \sigma^2 \mathbf{I}\right) \\
    & \boldsymbol{\mu}=\frac{\sqrt{\gamma_{t-1}}\left(1-\alpha_t\right)}{1-\gamma_t} \mathbf{Z}_0+\frac{\sqrt{\alpha_t}\left(1-\gamma_{t-1}\right)}{1-\gamma_t} \mathbf{Z}_t \\
    & \sigma^2=\frac{\left(1-\gamma_{t-1}\right)\left(1-\alpha_t\right)}{1-\gamma_t} .
    \end{aligned}
    \label{Equ:Forward-process-q-Z-t-1}
\end{equation}
% Since the conditional denoising Transformer predicts $\mathbf{Z}_0$, the posterior is useful in the reverse process.
This posterior is useful in the reverse process.

\vspace{-2.833mm}
% \subsubsection{Inference via Iterative Refinement}
\subsubsection{Reverse Markovian Process}
\vspace{-0.5mm}

The reverse process inferences $\mathbf{Z}_0$ via iterative refinements.
It starts from a pure Gaussian noise $\mathbf{Z}_T$ and goes in the opposite direction of the forward process:
\begin{equation}
    \begin{aligned}
    p_\theta\left(\mathbf{Z}_{0: T} \mid \mathbf{X}, \mathbf{Y}\right) & =p\left(\mathbf{Z}_T\right) \prod_{t=1}^T p_\theta\left(\mathbf{Z}_{t-1} \mid \mathbf{Z}_t, \mathbf{X}, \mathbf{Y}\right) \\
    p\left(\mathbf{Z}_T\right) & =\mathcal{N}\left(\mathbf{Z}_T ; \mathbf{0}, \mathbf{I}\right) \\
    p_\theta\left(\mathbf{Z}_{t-1} \mid \mathbf{Z}_t, \mathbf{X}, \mathbf{Y}\right) & =\mathcal{N}\left(\mathbf{Z}_{t-1} ; \mu_\theta\left(\mathbf{X}, \mathbf{Y}, \mathbf{Z}_t, \gamma_t\right), \sigma_t^2 \mathbf{I}\right),
    \end{aligned}
\end{equation}
where the distribution $p_\theta\left(\mathbf{Z}_{t-1} \mid \mathbf{Z}_t, \mathbf{X}, \mathbf{Y}\right)$ is parameterized with $\theta$.
Note that the CDFormer provides a prediction of $\hat{\mathbf{Z}}_0$.
% \begin{equation}
%     \hat{\mathbf{Z}}_0=\frac{1}{\sqrt{\gamma_t}}\left(\mathbf{Z}_t-\sqrt{1-\gamma_t} f_\theta\left(\boldsymbol{x}, \mathbf{Z}_t, \gamma_t\right)\right)
% \end{equation}
% \begin{equation}
%     \mu_\theta\left(\boldsymbol{x}, \mathbf{Z}_t, \gamma_t\right)=\frac{1}{\sqrt{\alpha_t}}\left(\mathbf{Z}_t-\frac{1-\alpha_t}{\sqrt{1-\gamma_t}} f_\theta\left(\boldsymbol{x}, \mathbf{Z}_t, \gamma_t\right)\right)
% \end{equation}
Thus, according to (\ref{Equ:Forward-process-q-Z-t-1}), each refinement step takes the following form:
% explain f_theta 
\begin{equation}
    \begin{aligned}
        \mathbf{Z}_{t-1} = &\frac{\sqrt{\gamma_{t-1}}\left(1-\alpha_t\right)}{1-\gamma_t} f_\theta\left(\mathbf{X}, \mathbf{Y}, \mathbf{Z}_t, \gamma_t\right)\\
        &+\frac{\sqrt{\alpha_t}\left(1-\gamma_{t-1}\right)}{1-\gamma_t} \mathbf{Z}_t + \sqrt{1-\alpha_t} \boldsymbol{\epsilon},
    \end{aligned}
\end{equation}
where $\boldsymbol{\epsilon} \sim \mathcal{N}\left(\mathbf{0},\mathbf{I}\right)$ and $f_{\theta}$ is the CDFormer.
% Conditional Denoising Transformer

\begin{figure*}
    \centering
    \includegraphics[width=0.9\linewidth]{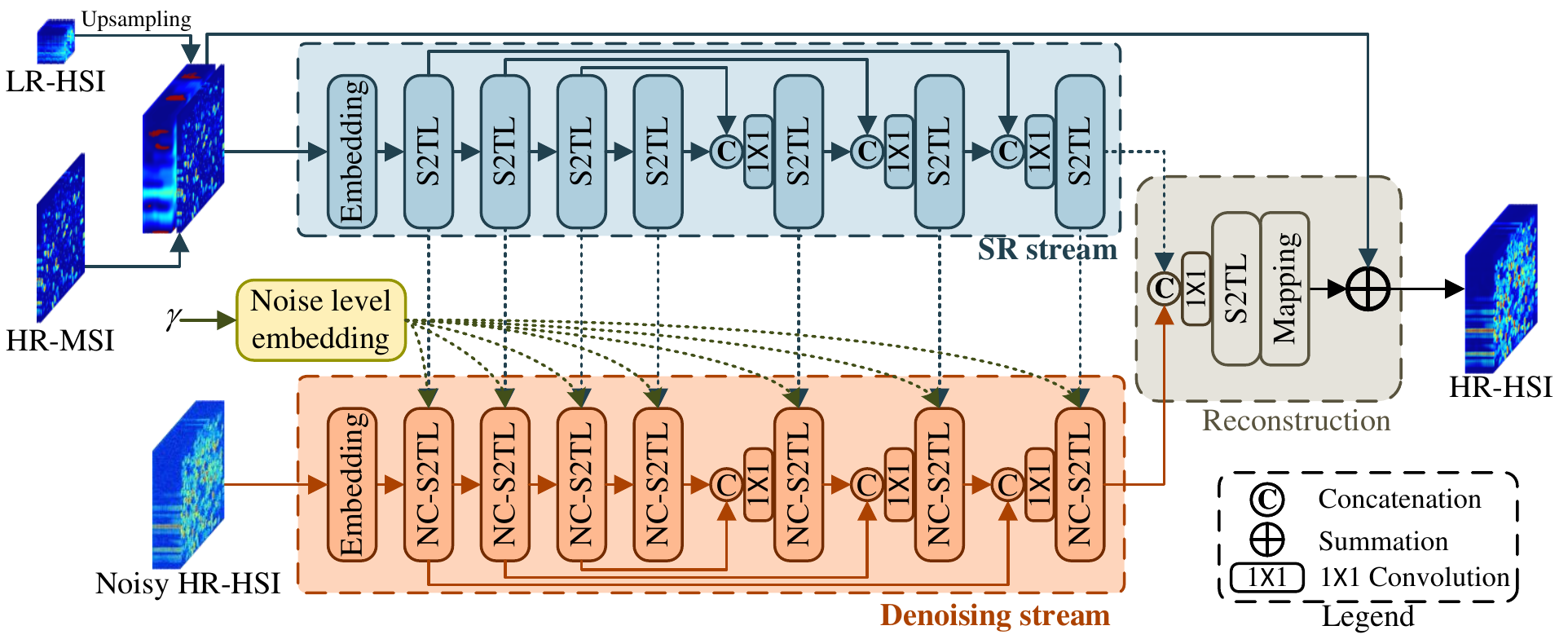}
    \caption{Architecture of Conditional Denoising Transformer.}
    \label{Fig:Conditional-denoising-Transformer}
    \vspace{-2.1mm}
\end{figure*}

\vspace{-2.833mm}
\subsubsection{Noise Schedule}
\label{Sec:Noise-schedule}
\vspace{-0.5mm}

Inspired by the research reported in \cite{chen2020wavegrad}, we sample $\gamma$ with two steps during training.
In particular, we first sample a time step $t \sim U\left\{1,T\right\}$ and then randomly select $\gamma \sim U\left(\gamma_{t-1},\gamma_t \right)$.
As such, $\gamma \sim p\left(\gamma\right)=\sum_{t=1}^{T}\frac{1}{T}U\left(\gamma_{t-1},\gamma_t\right)$.
Normally, the model with a large $T$ can achieve better results.
However, we find (through empirical investigations) that the performance is not very sensitive to the exact values of $T$.
Therefore, no hyper-parameter search about $T$ is conducted and we set $T=2000$ for simplicity.
As for the inference process, we set the maximum generation iterations to 100, employing a linear noise schedule.
% For our training noise schedule, we follow \cite{chen2020wavegrad}, and use a piece-wise distribution for $\gamma$, $p\left(\gamma\right)=\sum_{t=1}^{T}\frac{1}{T}U\left(\gamma_{t-1},\gamma_t\right)$.
% Specifically, during training, we first uniformly sample a time step $t\sim\left\{0,\ldots, T\right\}$ followed by sampling $\gamma\sim U\left(\gamma_{t-1},\gamma_t\right)$.
% For all experiments, we set $T=2000$ and the $\gamma_t$ are uniformly spaced.
% Larger values of $T$ generally yield better models, but model performance is relatively insensitive to the exact values of these parameters, so we do no hyper-parameter search during training.
% performing hyper-parameter search over the start and end noise levels.
% PSNR on held out data is used during the search to choose the best noise schedule.
% We also emphasize that this search is inexpensive as it does not require model retraining.

\vspace{-0.622mm}
\subsection{Conditional Denoising Transformer}
\vspace{-0.222mm}

The property of non-local self-similarity of HSIs is often exploited in denoising tasks but is usually not well captured by CNN-based models.
Due to the effectiveness of Transformer layer in capturing non-local long-range dependencies, the potential of Transformer is explored in conditional denoising of HSI.
% CDFormer is utilized to refine the noisy HR-HSI, conditioned on the feature representations of HR-MSI and LR-HSI. 
Unfortunately, the vanilla Transformer focuses only on spatial relationships between pixels while 
neglecting the spectral dimension.
%扩散模型中采用的降噪网络通常直接将噪声图像与the image as the condition，不利于提取LR-HSI和HR-MSI的空谱信息。
Besides, denoising networks in conditional diffusion models normally concatenate all images together as input, which may hinder the extraction of useful spatio-spectral information in LR-HSIs and HR-MSIs.
Hence, the CDFormer adopts a two-stream architecture and is constructed with stacked Spatio-Spectral Transformer Layers (S2TLs).

% CDFormer adopts a two-stream architecture including an SR stream and a denoising stream.
The architecture of the CDFormer is shown in Figure \ref{Fig:Conditional-denoising-Transformer}.
% The SR stream first utilizes a $3\times3$ convolution to extract low-level feature embeddings $\mathbf{F}_0^{SR}$ and then transform it into deep features $\mathbf{F}_l^{SR}$ with a stacked-S2TL encoder-decoder.
The SR stream first utilizes a $3\times3$ convolution to generate low-level feature embeddings $\mathbf{F}_0^{SR}$ and then transforms it into deep features $\mathbf{F}_l^{SR}$ with a stacked-S2TLs.
Instead of adapting $t$ as done in the existing work \cite{chen2020wavegrad}, our method is conditioned on $\gamma$ directly to achieve efficient generation.
% To condition the CDFormer on the information of the input image pair (LR-HSI and HR-MSI), we feed the feature map $\mathbf{F}_l^{SR}$ to the denoising stream at each level.
% The denoising stream is a symmetric sub-network containing multiple noise-aware conditional S2TLs (NC-S2TLs) that take as input the embedded noise level and the image representation $\mathbf{F}_l^{SR}$.
The denoising stream contains multiple noise-aware conditional S2TLs (NC-S2TLs) that take as input the embedded noise level and the image representation $\mathbf{F}^{SR}$.
The Reconstruction module is set to produce a noise-free HR-HSI, by employing residual learning to alleviate the difficulty of HR-HSI generation while mapping the features onto HR-HSI via a $3\times3$ convolution and addition operation.

\vspace{-2.833mm}
\subsubsection{Noise Level Embedding}
\vspace{-0.5mm}

The noise level offers essential information for denoising models.
Inspired by the work of \cite{chen2020wavegrad}, we embed noise level within the models with sinusoidal positional encoding.
The process of noise level embedding (NLE) can be formulated as follows:
\begin{equation}
    \begin{aligned}
        NLE_{\gamma, 2i}&=\sin \left(\gamma / 10000^{2i / C}\right) \\
        NLE_{\gamma, 2i+1}&=\cos \left(\gamma / 10000^{2i / C}\right),
    \end{aligned}
\end{equation}
where $C$ is the number of channels of S2TLs; $i\in[1,C/2]$.

\vspace{-2.833mm}
\subsubsection{Spatio-Spectral Transformer Layers}
\vspace{-0.5mm}

\begin{figure}
    \centering
    \includegraphics[width=0.85\linewidth]{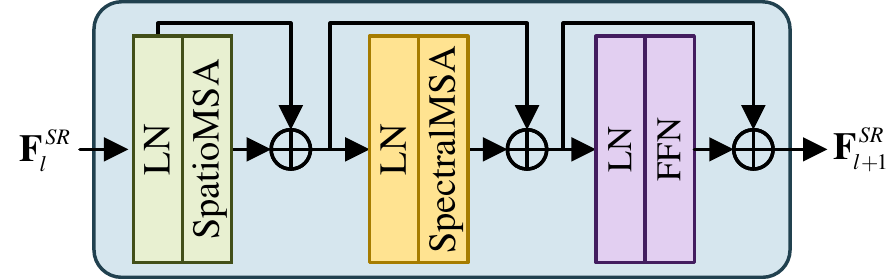} 
    \caption{Spatio-Spectral Transformer Layer, where ``LN" denotes layer normalization.}
    \label{Fig:S2TL}
    \vspace{-2.1mm}
\end{figure}

Figure \ref{Fig:S2TL} illustrates the architecture of one S2TL, which consists of a Spatial Multi-head Self-Attention (SpatioMSA), a Spectral Multi-head Self-Attention (SpectralMSA), and a Feed Forward Network (FFN).
SpatioMSA and SpectralMSA learn the interactions of spatial regions and inter-spectra relationships, respectively.
To alleviate the computational burden, we adopt the transposed attention \cite{zamir2022restormer} in SpectralMSA.
SpatioMSA applies the popular window partitioning strategy \cite{liu2021swin} to reduce the computational complexity.
In addition, the gating mechanism \cite{zamir2022restormer} is employed in the implementation of FFN.

\vspace{-2.833mm}
\subsubsection{Noise-Aware Conditional S2TLs}
\vspace{-0.5mm}

\begin{figure}
    \centering
    \includegraphics[width=0.68\linewidth]{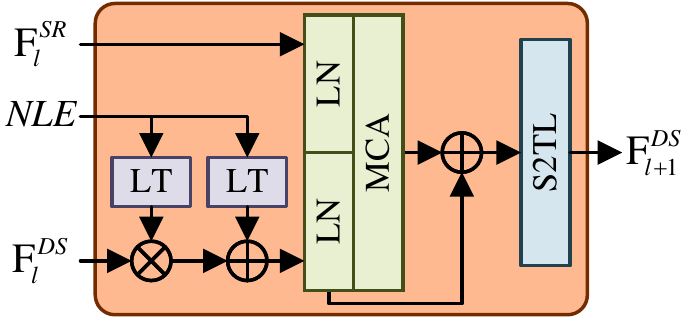} 
    \caption{Noise-Aware Conditional Spatio-Spectral Transformer Layer, where ``LT" represents linear transform, ``LN" denotes layer normalization, and ``MCA" is the abbreviation of multi-head cross attention.}
    \label{Fig:NC-S2TL}
    \vspace{-3mm}
\end{figure}

% \begin{figure}
%     \centering
%     \includegraphics[width=0.4\linewidth]{Figures/NC-S2TL.pdf}
%     \caption{Noise-Aware Conditional Spatio-Spectral Transformer Layer.}
%     \label{Fig:NC-S2TL}
% \end{figure}

To condition the overall model on the hierarchical features of SR stream, we feed $\mathbf{F}_l^{SR}$ to the Noise-Aware Conditional S2TLs (NC-S2TLs), each of which is a key building block of the denoising stream.
Figure \ref{Fig:NC-S2TL} depicts the structure of an NC-S2TL, which takes as input the $NLE$ (a vector), $\mathbf{F}_l^{SR}$ and $\mathbf{F}_l^{DS}$, where $\mathbf{F}_l^{SR}$ and $\mathbf{F}_l^{DS}$ have the same spatial resolution.
$NLE$ is first transformed and merged with $\mathbf{F}_l^{DS}$ with the result subsequently processed with the means of multi-head cross attention (MCA) \cite{chen2021crossvit} in order to condition the model on $\mathbf{F}_l^{SR}$.
As a result, each S2TL learns the spatio-spectral dependencies.

\vspace{-0.622mm}
% \subsection{Optimizing the Conditional Denoising Transformer with Progressive Learning}
\subsection{Progressive Learning}
\vspace{-0.222mm}

CNN-based restoration models are normally trained on fixed-size image patches.
However, training CDFormer on small cropped patches may not appropriately reflect the global image statistics, thereby providing suboptimal performance on full-resolution images when used.
To this end, we perform progressive learning where the network is trained on smaller image patches in the early epochs and on gradually larger patches in the later training epochs.
The model trained on mixed-size patches via progressive learning shows enhanced performance during testing where images can be of different resolutions (which is a common case in image restoration).
The progressive learning strategy behaves in a similar fashion to the curriculum learning process where the network starts with a simpler task and gradually moves to learning a more complex one (where the preservation of fine image structure/textures is required).

% Due to the limitation of GPU memory, we only train part of CDFormer on full-resolution.
To reduce the pressure on the demand of GPU memory, we only train the second half of CDFormer on full-resolution images.
% Since training on large patches comes at the cost of longer time, we reduce the batch size as the patch size increases to maintain a similar time per optimization step as of the fixed patch training.
The loss function used for such training is defined as follows:
% \begin{equation}
% \label{Equ:Loss-function-unsupervised-learning}
%     % \begin{aligned}
%         \mathcal{L}={\| {\mathbf{X}}-\mathbf{R}{\hat{\mathbf{Z}}} \|}_1+{\| {\mathbf{Y}}-{\hat{\mathbf{Z}}}\mathbf{D} \|}_1+ \|\mathbf{Z}-\hat{\mathbf{Z}}\|_1 ,
%     % \end{aligned},
%     % \vspace{-2mm}
% \end{equation}
\begin{equation}
\label{Equ:Loss-function-unsupervised-learning}
    \begin{aligned}
        \mathcal{L}&={\| {\mathbf{X}}-\mathbf{R}{\hat{\mathbf{Z}}_0} \|}_1+{\| {\mathbf{Y}}-{\hat{\mathbf{Z}}_0}\mathbf{D} \|}_1+ \|\mathbf{Z}_0-\hat{\mathbf{Z}}_0\|_1\\
        \hat{\mathbf{Z}}_0&=f_\theta(\sqrt{\gamma}\mathbf{Z}_0+\sqrt{1-\gamma}\boldsymbol{\epsilon},\mathbf{X},\mathbf{Y})
    \end{aligned},
    % \vspace{-2mm}
\end{equation}
where $\boldsymbol{\epsilon} \sim \mathcal{N}\left(\mathbf{0},\mathbf{I}\right)$,
% $\mathbf{Z}_0$ denotes the ground truth.
($\mathbf{X},\mathbf{Y},\mathbf{Z}$) is sampled from the training set,
% $\hat{\mathbf{Z}}$ is generated with CDFormer.
and the noise schedule about $\gamma$ has been discussed above.
The first two terms are designed according to the observation models, while the last one is based on the assumption of Laplace distribution.

\vspace{-0.811mm}
\section{Experiments}

% In this section, extensive experiments are conducted on three commonly-used public-available HSI-SR datasets to demonstrate the effectiveness of the proposed method.
Systematic experiments are herein conducted on four commonly-used public-available HSI-SR datasets to demonstrate the effectiveness of the proposed approach.

\vspace{-0.622mm}
\subsection{Datasets}
\vspace{-0.222mm}

Four datasets including CAVE \cite{yasuma2010generalized}, PaviaU \cite{2011PaviaUniversityScene}, Chikusei \cite{yokoya2016airborne}, and HypSen \cite{yang2018hyperspectral} are used in our experiments, with the following details on each.

\vspace{-2.444mm}
\paragraph{CAVE:}
There are 32 scenes with a spatial size of $512\times512$ in the CAVE dataset, where we select the first 20 HSIs for training, with the remaining 12 images used for testing.
We generate LR-HSIs by Gaussian blur and down-sampling using a factor of 32 as done in \cite{xie2020mhf}.
HR-MSIs are acquired by integrating all HR-HSI bands according to the spectral response function of Nikon D700.
The original HR-HSIs are treated as ground truth.

\vspace{-2.444mm}
\paragraph{PaviaU:}
Collected by the University of Pavia, Italy, the original HSI dataset consists of $610\times340$ pixels in which the top-left $128\times128$ area is extracted as the test data, with the remaining used for training.
Except for water absorption bands, all other 103 bands are chosen for the experiments.
Note that the down-sampling factor for the generation of LR-HSIs is four, and the spectral response function is the same as that of the WorldView-3 satellite. 

\vspace{-2.444mm}
\paragraph{Chikusei:}
This dataset consists of 128 bands with a spectral range of 363$nm$ to 1018$nm$ and a spatial resolution of $2517\times2335$.
The original HSI data was taken by an airborne visible and near-infrared imaging sensor over Chikusei, Japan.
To alleviate the impact of the back boundary and noise, we crop the center area and remove noise bands.
The processed image has a size of $2048 \times 2048 \times 110$.
The top half $1024 \times 2048 \times 110$ area is selected as the training data, while the rest half is split into eight testing $512 \times 512 \times 110$ patches.
For the production of LR-HSIs and HR-MSIs, this dataset adopts the same processing as with PaviaU.

\begin{figure*}[!t]
    \centering
    % \vspace{-5mm}
    % \hspace{-6.55mm}
    \includegraphics[width=0.133\linewidth]{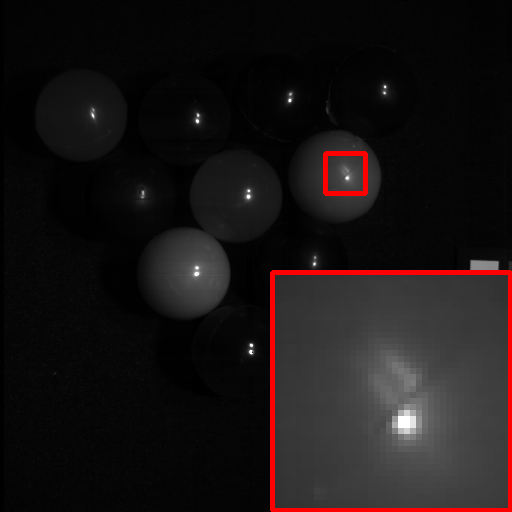}\hspace{0.881mm}
    \includegraphics[width=0.133\linewidth]{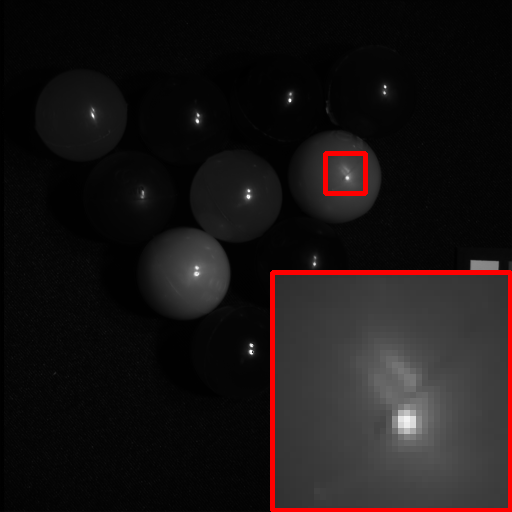}\hspace{0.881mm}
    \includegraphics[width=0.133\linewidth]{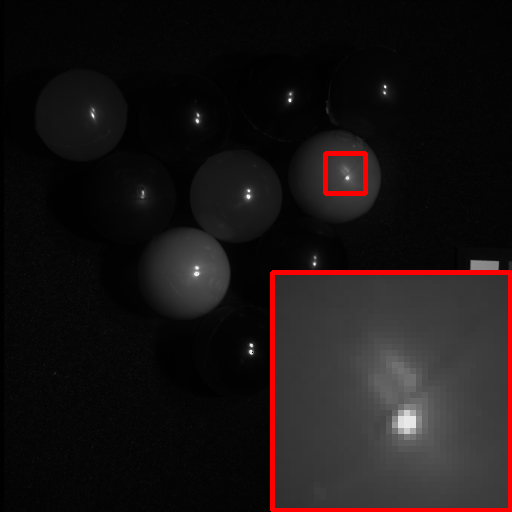}\hspace{0.881mm}
    \includegraphics[width=0.133\linewidth]{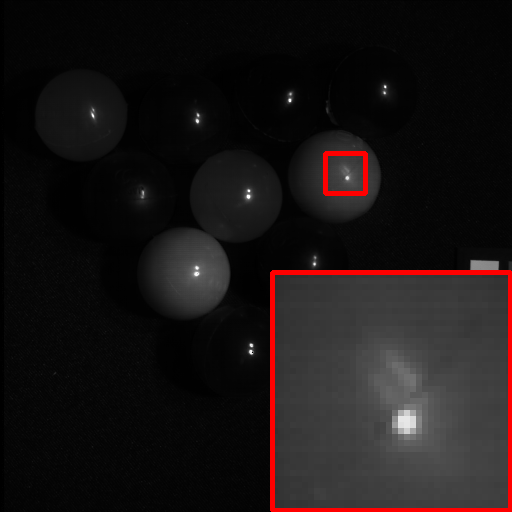}\hspace{0.881mm}
    \includegraphics[width=0.133\linewidth]{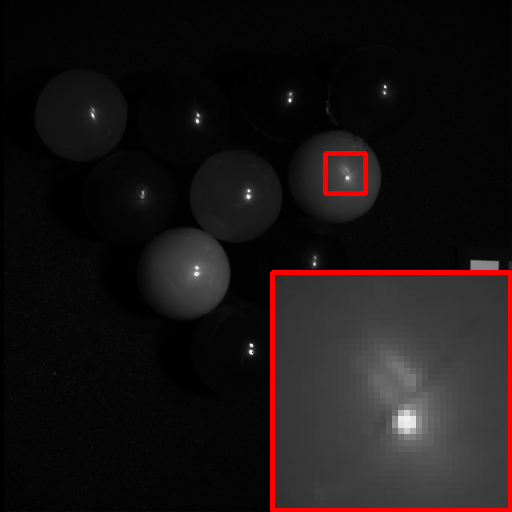}\hspace{0.881mm}
    \includegraphics[width=0.133\linewidth]{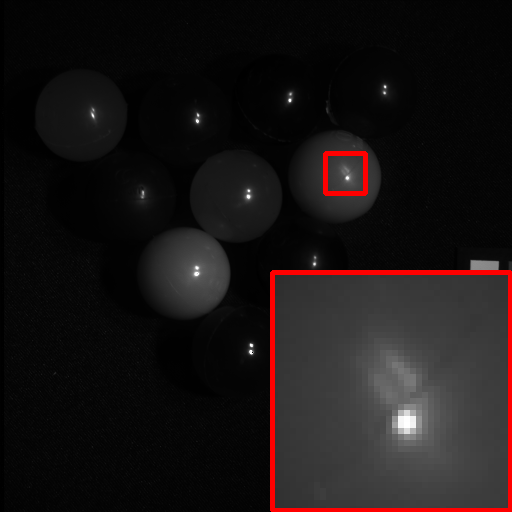}\hspace{0.881mm}
     \includegraphics[width=0.133\linewidth]{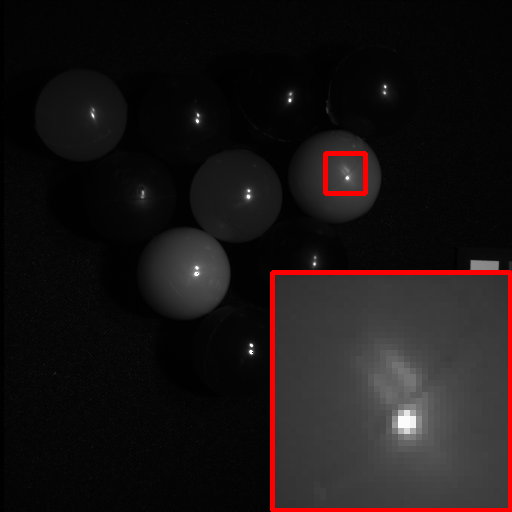}
    
    \vspace{-3.5mm}
    \subfloat[UTV-TD]{
        \includegraphics[width=0.133\linewidth]{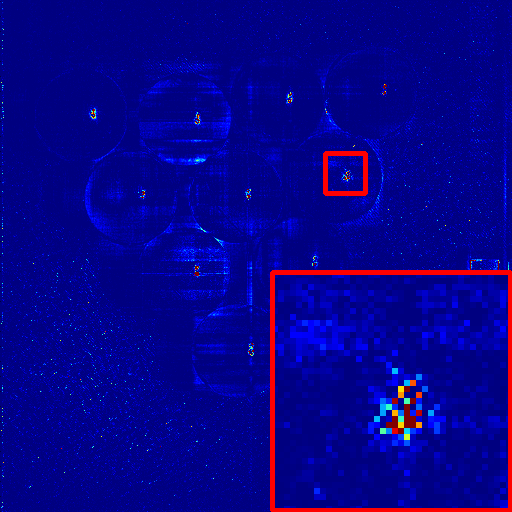}
    }
    \subfloat[UAL]{
        \includegraphics[width=0.133\linewidth]{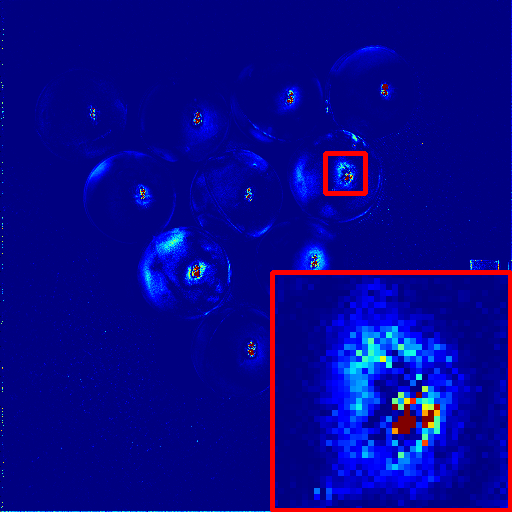}
    }
    \subfloat[BRResNet]{
        \includegraphics[width=0.133\linewidth]{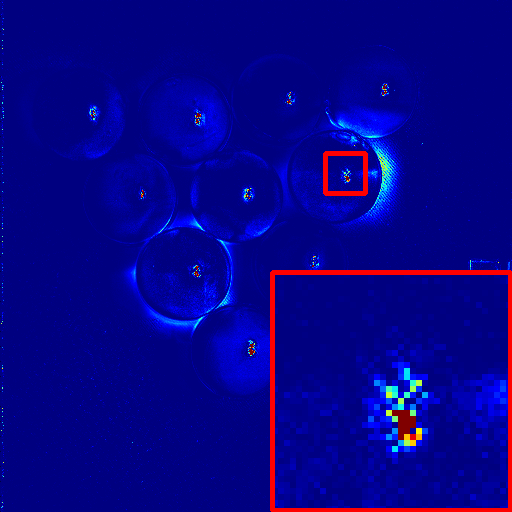}
    }
    \subfloat[CMHF-Net]{
        \includegraphics[width=0.133\linewidth]{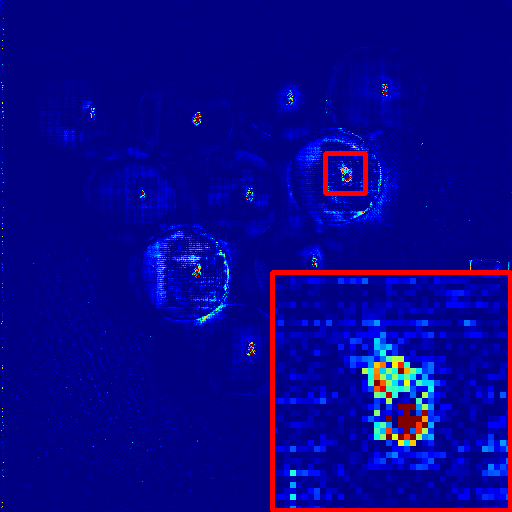}
    }    
    \subfloat[UAL-DMI]{
        \includegraphics[width=0.133\linewidth]{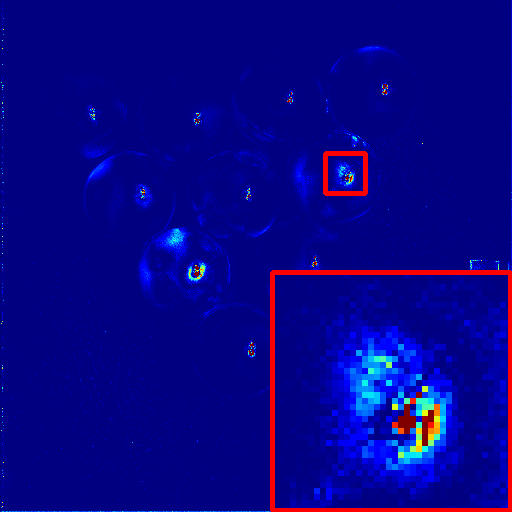}
    }
    \subfloat[HSR-Diff]{
        \includegraphics[width=0.133\linewidth]{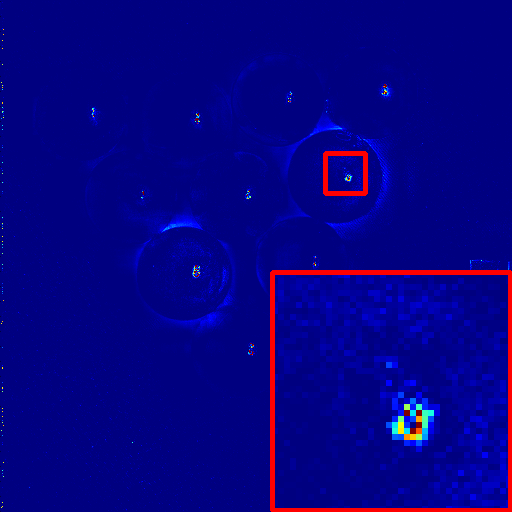}
    }
    \subfloat[Ground truth]{
        \includegraphics[width=0.133\linewidth]{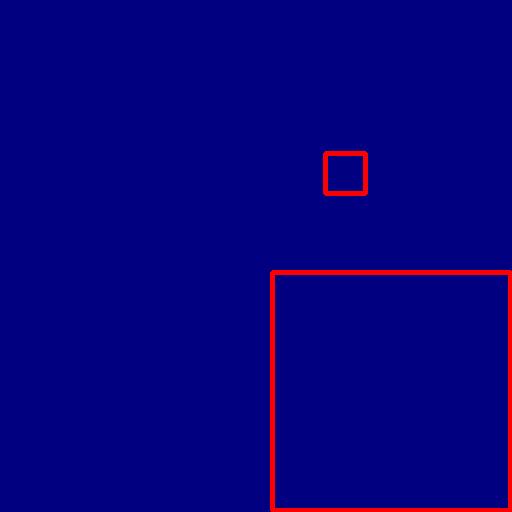}
    }
    
    \includegraphics[width=0.3\linewidth]{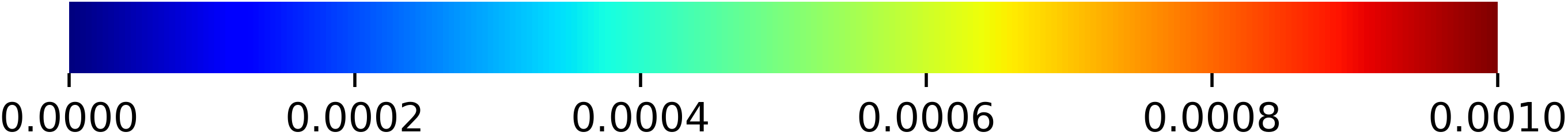}
    \vspace{-3mm}
    \caption{Visual quality comparison for fused HSIs of all competing methods on CAVE, where first and second rows show fourth band and corresponding heatmaps (mean squared error), respectively.}
    % \vspace{-3mm}
    \label{Fig:Visualized-results-CAVE}
    \vspace{-2.1mm}
\end{figure*}

\begin{figure*}[!t]
    \centering
    % \vspace{-5mm}
    % \hspace{-6.55mm}    
    \includegraphics[width=0.133\linewidth]{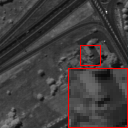}\hspace{0.881mm}
    \includegraphics[width=0.133\linewidth]{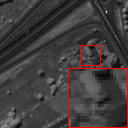}\hspace{0.881mm}
    \includegraphics[width=0.133\linewidth]{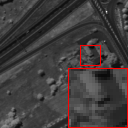}\hspace{0.881mm}
    \includegraphics[width=0.133\linewidth]{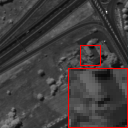}\hspace{0.881mm}
    \includegraphics[width=0.133\linewidth]{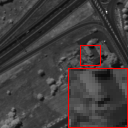}\hspace{0.881mm}
    \includegraphics[width=0.133\linewidth]{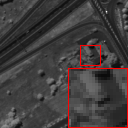}\hspace{0.881mm}
     \includegraphics[width=0.133\linewidth]{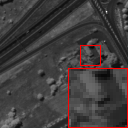}
	
    \vspace{-3.5mm}
    \subfloat[UTV-TD]{
        \includegraphics[width=0.133\linewidth]{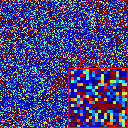}
    }
    \subfloat[UAL]{
        \includegraphics[width=0.133\linewidth]{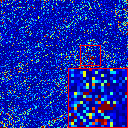}
    }
    \subfloat[BRResNet]{
        \includegraphics[width=0.133\linewidth]{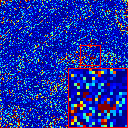}
    }
    \subfloat[CMHF-Net]{
        \includegraphics[width=0.133\linewidth]{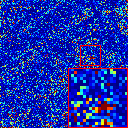}
    }
    \subfloat[UAL-DMI]{
        \includegraphics[width=0.133\linewidth]{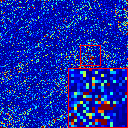}
    }
    \subfloat[HSR-Diff]{
        \includegraphics[width=0.133\linewidth]{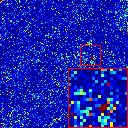}
    }	
    \subfloat[Ground truth]{
        \includegraphics[width=0.133\linewidth]{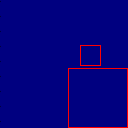}
    }
    
    \includegraphics[width=0.3\linewidth]{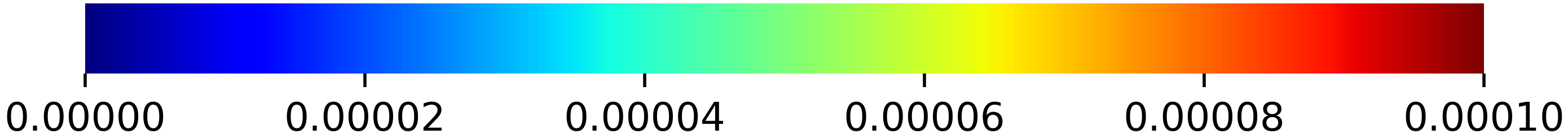}
    \vspace{-3mm}
    \caption{Visual quality comparison for fused HSIs of all competing methods on PaviaU, where first and second rows show 81st band and corresponding heatmaps (mean squared error), respectively.
    }
    % \vspace{-3mm}
    \label{Fig:Visualized-results-PaviaU}
    \vspace{-2.1mm}
\end{figure*}

\begin{figure*}[!t]
    \centering
    % \vspace{-5mm}
    % \hspace{0.07mm}
    \includegraphics[width=0.133\linewidth]{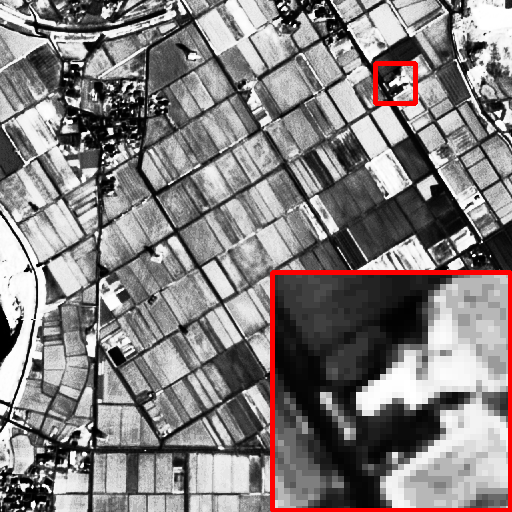}\hspace{0.881mm}
    \includegraphics[width=0.133\linewidth]{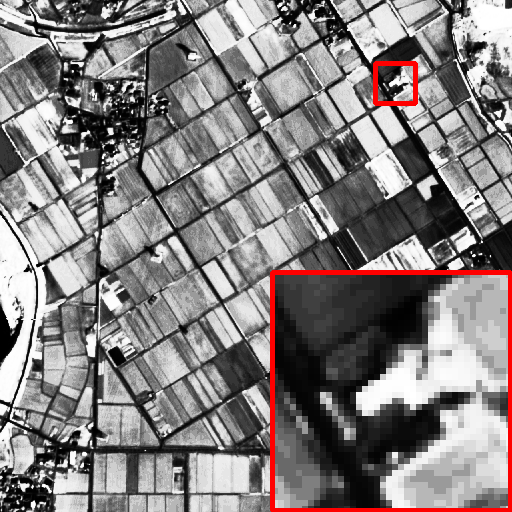}\hspace{0.881mm}
    \includegraphics[width=0.133\linewidth]{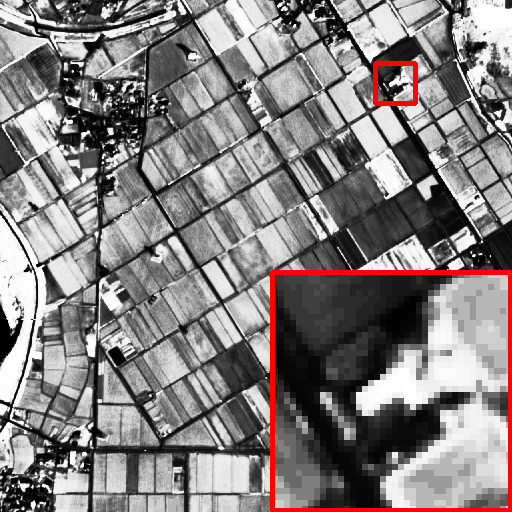}\hspace{0.881mm}
    \includegraphics[width=0.133\linewidth]{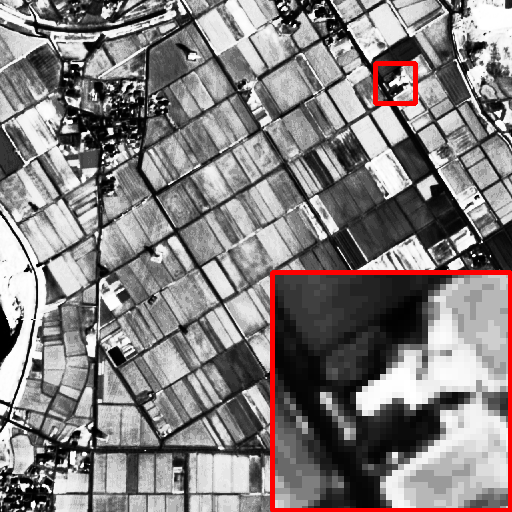}\hspace{0.881mm}
    \includegraphics[width=0.133\linewidth]{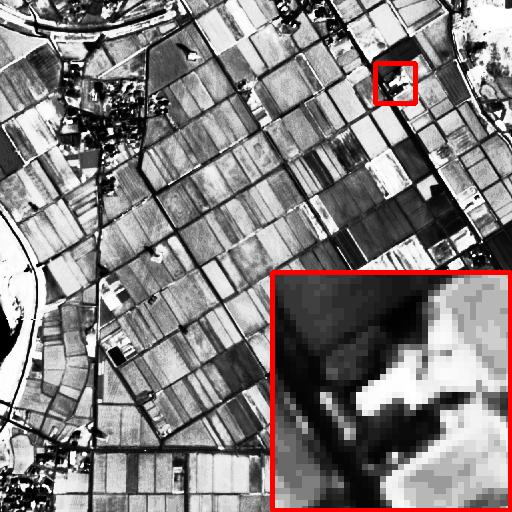}\hspace{0.881mm}
    \includegraphics[width=0.133\linewidth]{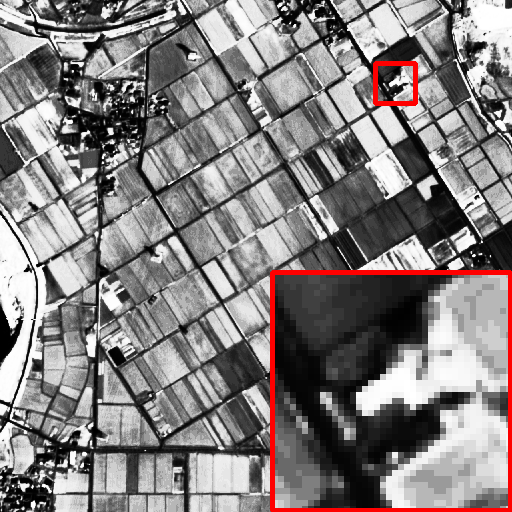}\hspace{0.881mm}
    \includegraphics[width=0.133\linewidth]{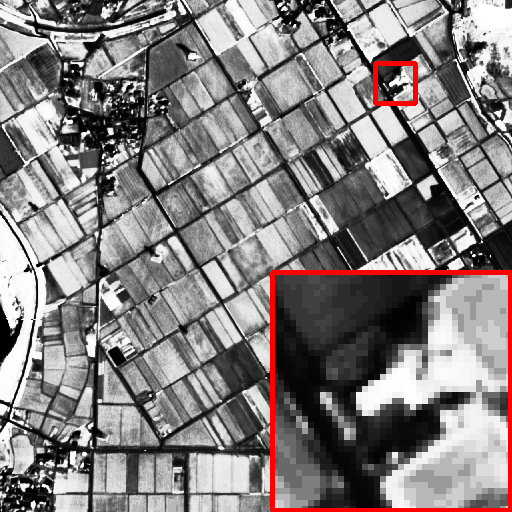}
	
    \vspace{-3.5mm}
    \subfloat[UTV-TD]{
        \includegraphics[width=0.133\linewidth]{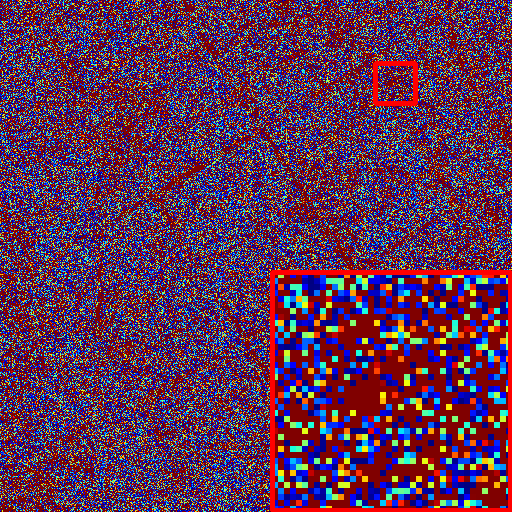}
    }
    \subfloat[UAL]{
        \includegraphics[width=0.133\linewidth]{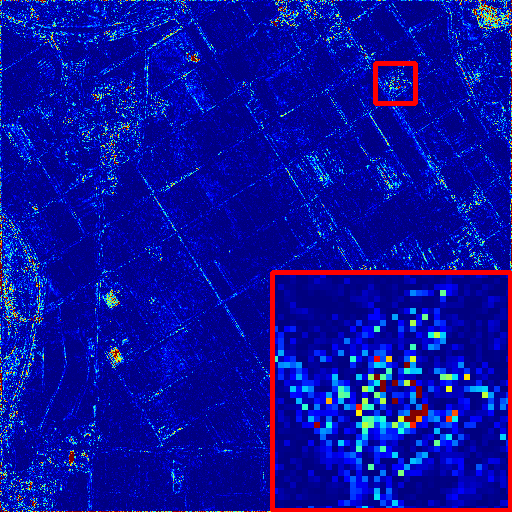}
    }
    \subfloat[BRResNet]{
        \includegraphics[width=0.133\linewidth]{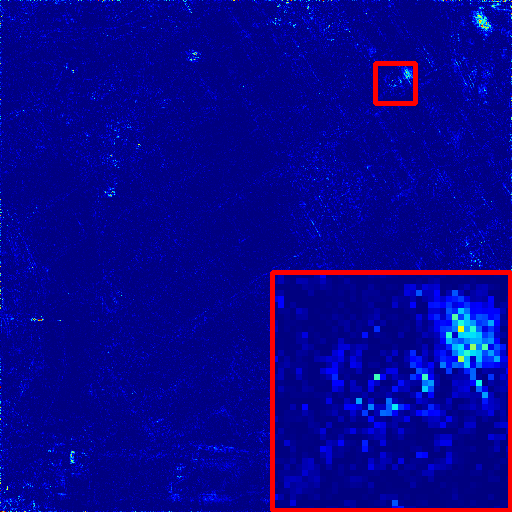}
    }
    \subfloat[CMHF-Net]{
        \includegraphics[width=0.133\linewidth]{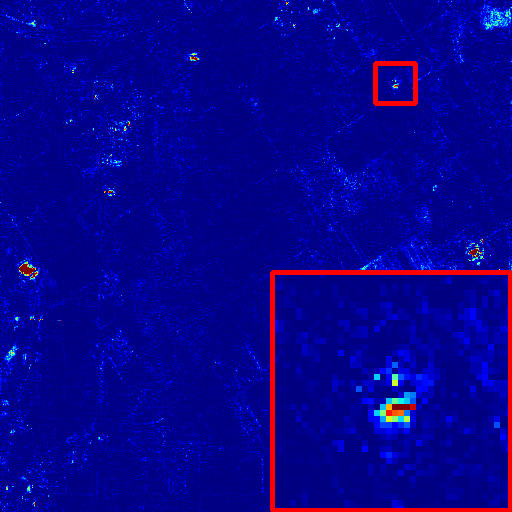}
    }
    \subfloat[UAL-DMI]{
        \includegraphics[width=0.133\linewidth]{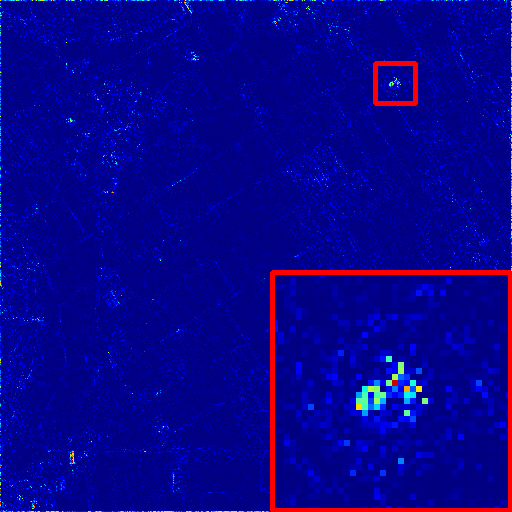}
    }
    \subfloat[HSR-Diff]{
        \includegraphics[width=0.133\linewidth]{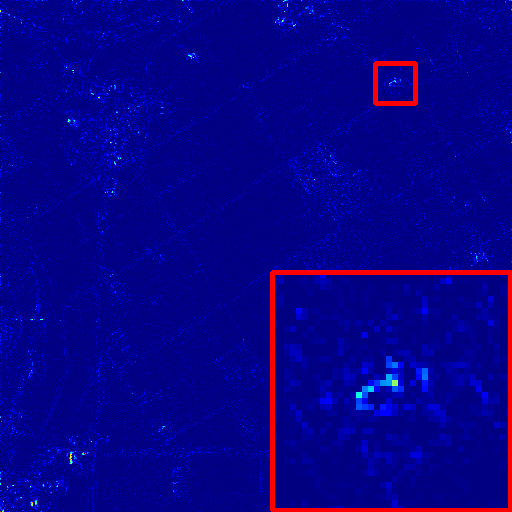}
    }	
    \subfloat[Ground truth]{
        \includegraphics[width=0.133\linewidth]{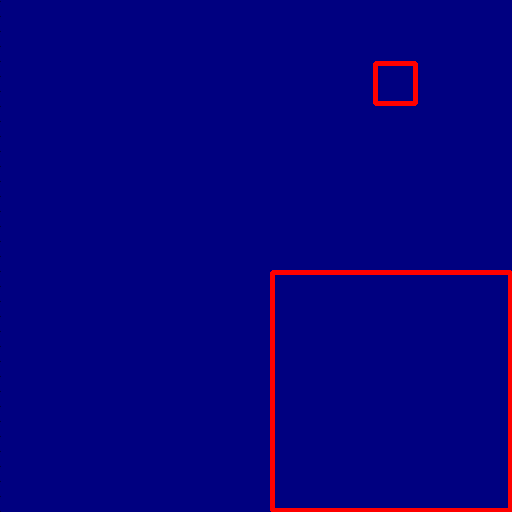}
    }
    
    \includegraphics[width=0.3\linewidth]{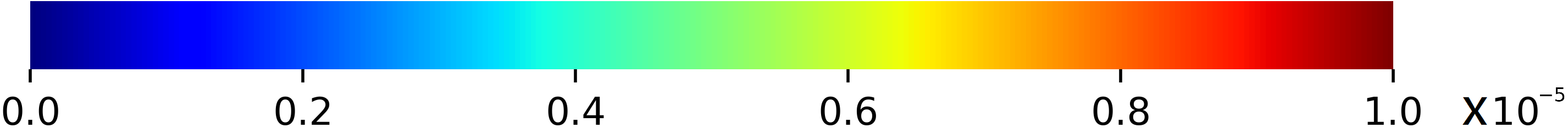}
 
    \vspace{-3mm}
    \caption{Visual quality comparison for fused HSIs of all competing methods on Chikusei, where first and second rows show 67th band and corresponding heatmaps (mean squared error), respectively.
    }
    % \vspace{-3mm}
    \label{Fig:Visualized-results-Chikusei}
    \vspace{-3.1mm}
\end{figure*}

\vspace{-2.444mm}
\paragraph{HypSen:}
This dataset concerns a real scenario consisting of a 30m-resolution HSI and a 10m-resolution MSI. 
The Hyperion sensor on the Earth Observing-1 satellite provided the HSI with 242 spectral bands in the spectral range of 400 2500nm, and the MSI with 13 bands was captured by the Sentinel-2A satellite. 
The blue, green, red, and near-infrared bands of MSI in our experiments are selected due to their high spatial resolution. 
To eliminate the impact of noise and water absorption, we remove those relevant bands, with 84 bands remaining in the HSI. 
We crop sub-images of size 250$\times$330 and 750$\times$990 from the Hyperion HSI and Sentinel-2A MSI respectively, in our study, with the pairs of sub-image patches spatially registered.

\vspace{-0.622mm}
\subsection{Methods Compared and Evaluation Metrics Used}
\vspace{-0.222mm}

Five state-of-the-art HSI-SR approaches are taken for comparison, including:  UTV-TD \cite{xu2020hyperspectral}, UAL \cite{zhang2020unsupervised}, BRResNet \cite{jin2021bam}, CMHF-Net \cite{xie2020mhf}, and UAL-DMI \cite{wang2022hyperspectral}.
% UTV-TD is a tensor-based approach.
% UAL, BRResNet, CMHF-Net fall into the category of the DL-based methods.
% UAL-DMI can be regarded as an upgraded version of UAL.
UTV-TD is a tensor-based technique; UAL, BRResNet, CMHF-Net fall into the category of the DL-based methods; and UAL-DMI can be regarded as an upgraded version of UAL.

Four quantitative quality metrics are employed for performance evaluation, including peak signal-to-noise ratio (PSNR), spectral angle mapper (SAM), erreur relative globale adimensionnelle de synthèse (ERGAS, namely error relative global dimensionless synthesis), and structure similarity (SSIM).
The smaller ERGAS and SAM are, the larger PSNR and SSIM are, and the better the fusion result is.

\vspace{-0.622mm}
\subsection{ Implementation Specification}
\vspace{-0.222mm}

% The batch size is set to 1.
% The learning rate is initialized as $1\times10^{-5}$ and decays to $1\times10^{-6}$ at 20 epochs.
% The weight decay is set as $1\times10^{-4}$.
% The weights of learnable layers are initialized by Kaiming’s initialization \cite{he2015delving}.
% The hyperparameters are determined via grid search.
All DL-based methods are trained on the same datasets.
For those compared methods, we use the publicly available source codes with default hyper-parameters as given in the corresponding research papers.
Our HSR-Diff is implemented on the PyTorch framework.
The learnable parameters of the CDFormer are initialized with Kaiming initialization \cite{he2015delving} and trained on 2 NVIDIA GeForce GTX 3090s.
The number of its channels is set to 256.
We utilize the Adam optimizer with $\beta_1=0.9$ and $\beta_2=0.999$ to optimize the CDFormer.
% With the limited GPU memory, the batch size for training $64 \times 64$ resolutions 
% the batch size is set as 2 at $128\times 128$ and $256 \times 256$ resolutions.
With limited GPU memory, the batch size is set to 4 and 2 for $128^2$ and $512^2$ images, respectively.
% the batch size is set as 2 for  $256 \times 256$ and $512 \times 512$ images
% With the limited GPU memory, the batch size is set as 2 at $128\times 128$ and $256 \times 256$ resolutions.
It costs 20000 epochs on the CAVE and PaviaU datasets while consuming 5000 epochs on the Chikusei dataset.
The learning rate is set as $1\times10^{-4}$.

\vspace{-0.622mm}
\subsection{Comparisons with State-of-the-art Methods}
\vspace{-0.222mm}

In this set of experiments, the evaluations are carried out using the first three datasets listed above without involving the real-world dataset, HypSen (which will be dealt with in the next section).

\vspace{-2.444mm}
\paragraph{Qualitative Comparison.}
To assess the performance of HSR-Diff qualitatively, we visualize example bands of HSIs in Figures. \ref{Fig:Visualized-results-CAVE}, \ref{Fig:Visualized-results-PaviaU}, and \ref{Fig:Visualized-results-Chikusei}.
It can be seen from these visual results that all compared methods produce satisfactory outcomes.
In particular, HSR-Diff generates gives the best result with minor errors since the corresponding MSE (mean squared error) images are much clearer than the others.

\vspace{-2.444mm}
\paragraph{Quantitative Comparison.}
To further verify the superior performance of the proposed HSR-Diff, quantitative results are presented in Table \ref{Tab:Quantative-results}.
Note that the performance indices on the CAVE and Chikusei datasets are averaged over all testing samples (12 samples for CAVE and eight samples for Chikusei), respectively.
It can be inferred from the results that the proposed HSR-Diff surpasses all competitors with a clear margin on all evaluation metrics.

\begin{table}
    \centering
    % \vspace{-2 mm}
    % \renewcommand{\arraystretch}{1.3} %调行距
    \setlength\tabcolsep{2pt} %调列距
    \begin{tabular}{c l r r r r}
        \toprule
        Dataset  & Methods & PSNR $\uparrow$  & SSIM $\uparrow$& SAM $\downarrow$ & ERGAS $\downarrow$\\
        \hline
        % \multirow{9}{*}{CAVE} & NLSTF\_SMBF \cite{dian2019nonlocal} & 37.66 & 0.9837 & 10.29 & 0.366\\
                                & UTV-TD     & 38.66 & 0.9799 & 7.98 & 0.329\\
                                & UAL        & 40.55 & 0.9933 & 4.33 & 0.271\\
                 CAVE           & BRResNet   & 41.36 & 0.9929 & 4.70 & 0.250\\
              32$\times$        & CMHF-Net   & 42.54 & 0.9939 & 4.69 & 0.216\\
                                & UAL-DMI    & 42.74 & 0.9950 & 3.79 & 0.213\\
                                & HSR-Diff       & \textbf{44.33} & \textbf{0.9951} & \textbf{3.71} & \textbf{0.179}\\
        \hline
                                & UTV-TD     & 44.46 & 0.9952 & 1.80 & 1.236\\
                                & UAL        & 45.42 & 0.9964 & 1.54 & 1.148\\
                PaviaU          & BRResNet   & 45.53 & 0.9965 & 1.53 & 1.111\\
               4$\times$        & CMHF-Net   & 45.77 & 0.9965 & 1.50 & 1.096\\
                                & UAL-DMI    & 45.68 & 0.9966 & 1.49 & 1.113\\
                                & HSR-Diff       & \textbf{46.47} & \textbf{0.9977} & \textbf{1.45} & \textbf{1.053}\\
        \hline
                                    & UTV-TD    & 48.38 & 0.9989 & 0.99  & 1.303\\
                                    & UAL       & 56.18 & 0.9998 & 0.49  & 0.421\\
             Chikusei               & BRResNet  & 56.79 & 0.9998 & 0.46  & 0.366\\
             4$\times$              & CMHF-Net  & 55.99 & 0.9998 & 0.50  & 0.483\\
                                    & UAL-DMI   & 56.57 & 0.9998 & 0.47  & 0.387\\
                                    & HSR-Diff      & \textbf{57.34} & \textbf{0.9999} & \textbf{0.43}  & \textbf{0.324}\\
        \bottomrule
    \end{tabular}
    \caption{Averaged PSNR, SSIM, SAM, and ERGAS of compared methods on CAVE, PaviaU, and Chikusei datasets.}
    \vspace{-3mm}
    % \vspace{-2.1mm}
    \label{Tab:Quantative-results}
\end{table}

\vspace{-0.622mm}
\subsection{Ablation Study}
\vspace{-0.222mm}

\begin{figure*}[!htbp]
    \centering
    % \vspace{-3.5mm}
    \subfloat[HSI]{
        \includegraphics[width=0.133\linewidth]{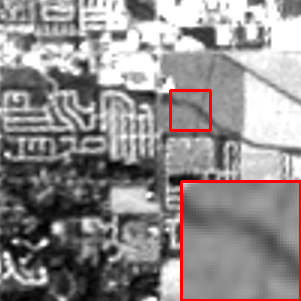}
    }
    \subfloat[UTV-TD]{
        \includegraphics[width=0.133\linewidth]{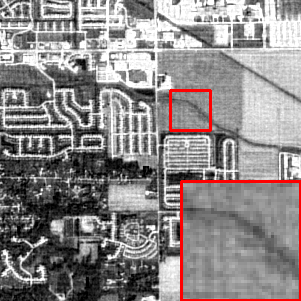}
    }
    \subfloat[UAL]{
        \includegraphics[width=0.133\linewidth]{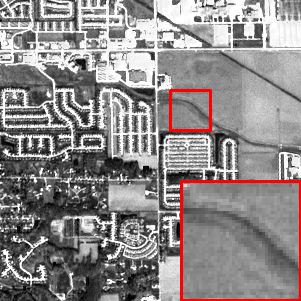}
    }
    \subfloat[BRResNet]{
        \includegraphics[width=0.133\linewidth]{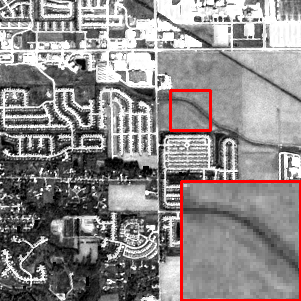}
    }
    \subfloat[CMHF-Net]{
        \includegraphics[width=0.133\linewidth]{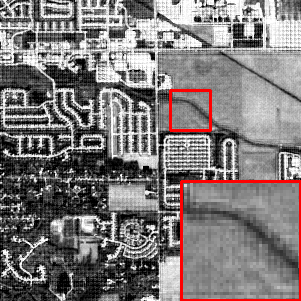}
    }    
    \subfloat[UAL-DMI]{
        \includegraphics[width=0.133\linewidth]{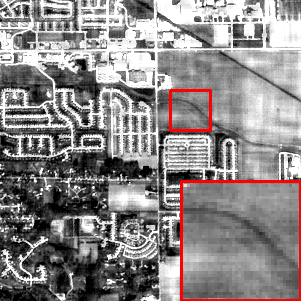}
    }
    \subfloat[HSR-Diff]{
        \includegraphics[width=0.133\linewidth]{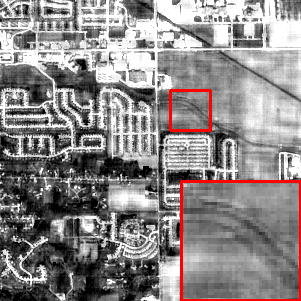}
    }	
    \vspace{-3mm}
    \caption{Visual fusion results of all competing methods for HypSen.
    }
    \vspace{-2mm}
    \label{Fig:Visualized-results-HypSen}
    \vspace{-2.1mm}
\end{figure*}

% \vspace{-2.444mm}
% Diffusion model: loss function (predict noise or HSI), diffusion model vs regression model.
\paragraph{Effect of conditional diffusion models.}

\begin{figure}
    \centering
    \includegraphics[width=0.3\linewidth]{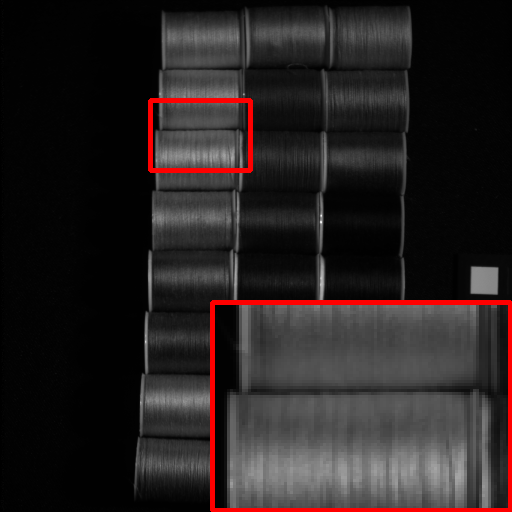}\hspace{0.881mm}
    \includegraphics[width=0.3\linewidth]{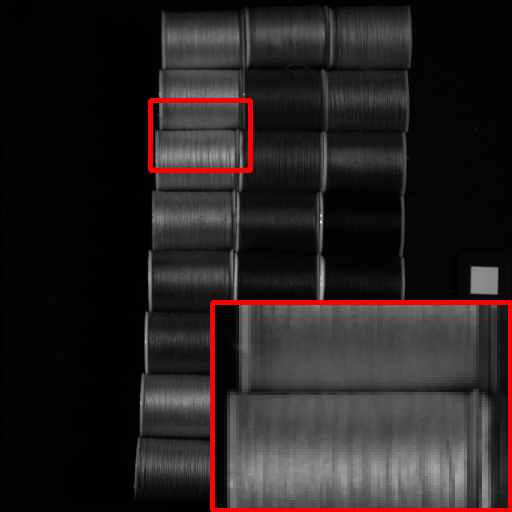}\hspace{0.881mm}
    \includegraphics[width=0.3\linewidth]{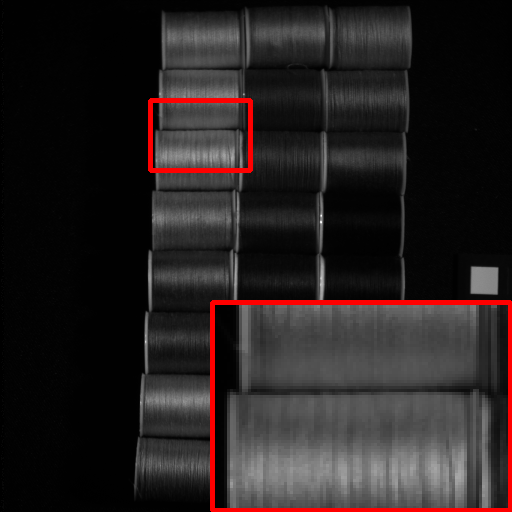}
 
    \vspace{-3.5mm}
    \subfloat[Regression]{
        \includegraphics[width=0.3\linewidth]{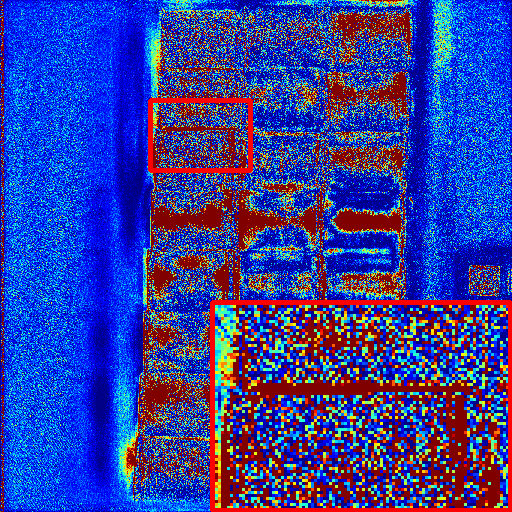}
    }
    \subfloat[HSR-Diff]{
        \includegraphics[width=0.3\linewidth]{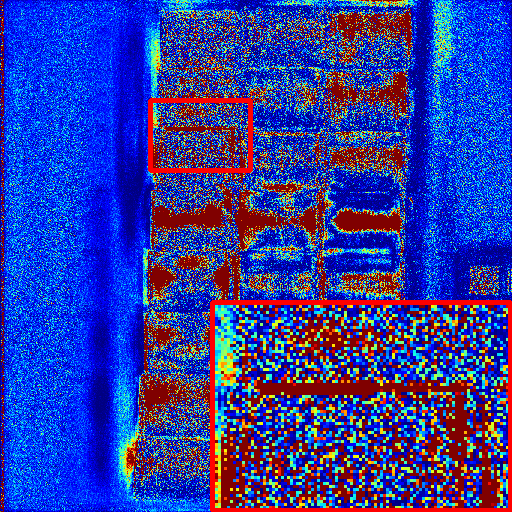}
    }
    \subfloat[Ground truth]{
        \includegraphics[width=0.3\linewidth]{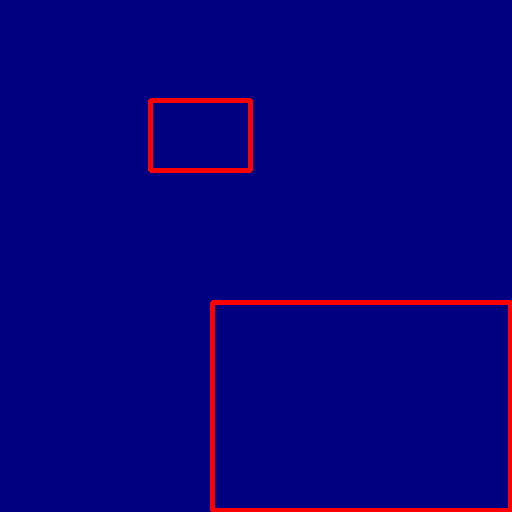}
    }
    
    \includegraphics[width=0.5\linewidth]{Results/heat-map-horizontal-bar-Chikusei.png}
    \vspace{-2mm}
    \caption{Fusion results (32$\times$) for HSR-Diff and Regression on the \textit{thread spools} image of CAVE.}
    \label{Fig:Ablation-study-Diffusion-vs-Regression}
    \vspace{-2.1mm}
\end{figure}

% Much of the early works on HSI-SR are based on the regression model.
% % To demonstrate the effect of the diffusion model, we train the regression model with the CDFormer.
% To compare the effect of the diffusion model and regression model, we train the regression model with the CDFormer.
% Note that the loss function, optimizer, and hyper-parameters are the same as the diffusion model.
% % Figure \ref{Fig:Ablation-study-Diffusion-vs-Regression} compares the HSR-Diff and Regression model, where the outputs of the Regression model are relatively blurry and lack fine-grained structure.
% Figure \ref{Fig:Ablation-study-Diffusion-vs-Regression} presents the fused results and corresponding error maps of HSR-Diff and regression model.
% As can be seen from the error maps, the HSI produced by HSR-Diff has less distortion than that of regression model.
% HSR-Diff relies on a series of iterative refinement steps, which permits our iterative approach to capture richer distribution.

Much of the early work on HSI-SR was based on the use of regression models.
To compare the effects of the diffusion and regression models, we train regression models containing the CDFormer.
Note that the loss function, optimizer, and hyper-parameters are all the same as the conditional diffusion models.
Figure \ref{Fig:Ablation-study-Diffusion-vs-Regression} presents the fused results and corresponding error maps of utilising HSR-Diff and regression models.
As can be seen from the error maps, the HSIs produced by HSR-Diff have less distortion than those by the regression models. This is because HSR-Diff works with a series of iterative refinement steps, facilitating the capture of richer information on data distributions of HR-HSIs.

\vspace{-2.444mm}
\paragraph{Effect of CDFormer.}

\begin{table}
    \centering
    % \vspace{-2 mm}
    % \renewcommand{\arraystretch}{1.3} %调行距
    \setlength\tabcolsep{2pt} %调列距
    \begin{tabular}{c l r r r r}
        \toprule
        Dataset  & Network & PSNR $\uparrow$  & SSIM $\uparrow$& SAM $\downarrow$ & ERGAS $\downarrow$\\
        \hline
        % \multirow{9}{*}{CAVE} & NLSTF\_SMBF \cite{dian2019nonlocal} & 37.66 & 0.9837 & 10.29 & 0.366\\
        \multirow{3}{*}{\makecell[c]{CAVE\\32$\times$}}   & U-Net   & 38.84  & 0.9797 & 7.32 & 0.318 \\
                                & C-w/o-SR    & 43.74 & 0.9942 & 3.94 & 0.188 \\
                                & CDFormer       & \textbf{44.33} & \textbf{0.9951} & \textbf{3.71} & \textbf{0.179}\\
        \hline
        \multirow{3}{*}{\makecell[c]{PaviaU\\4$\times$}}               & U-Net   & 42.75 & 0.9962 & 1.75 & 1.362 \\
                                & C-w/o-SR    & 46.08 & 0.9976 & 1.47 & 1.080\\
                                & CDFormer       & \textbf{46.47} & \textbf{0.9977} & \textbf{1.45} & \textbf{1.053}\\
        \hline
        \multirow{3}{*}{\makecell[c]{Chikusei\\4$\times$}}                   & U-Net  & 47.63 & 0.9980 & 1.20  & 1.794  \\
                                    & C-w/o-SR   & 56.68 & \textbf{0.9999} & 0.46  & 0.425 \\
                                    & CDFormer     & \textbf{57.34} & \textbf{0.9999} & \textbf{0.43}  & \textbf{0.324}\\
        \bottomrule
    \end{tabular}
    \caption{Ablation study on CDFormer.}
    \vspace{-3mm}
    \label{Tab:Ablation-study-CDFormer}
    % \vspace{-2.1mm}
\end{table}

Recall that CDFormer is conditioned on the hierarchical representations of HR-MSI and LR-HSI via a two-stream architecture.
However, alternative outstanding diffusion models \cite{ho2020denoising,saharia2022image} that are also excellent for conditional image generation are equipped with CNN-based U-Nets, where degenerated images are concatenated with noisy high-resolution output images.
To show the effectiveness of hierarchical representations, we remove the SR stream of CDFormer and name the resulting network ``C-w/o-SR".
In addition, we compare CDFormer with a CNN version of CDFormer that replaces all S2TL  with convolutional layers and show its results as ``CDCNN" in Table \ref{Tab:Ablation-study-CDFormer}.
The quantitative results show the use of CDFormer performs better than CDCNN, demonstrating the effectiveness of global statistics.
Indeed, with the two-stream architecture, CDFormer offers the best results thanks to the use of hierarchical features.

% CDFormer: vs U-Net; condition on images (concatenation of input images) vs hierarchical features; 
\vspace{-2.444mm}
\paragraph{Effect of progressive learning.}

\begin{table}
    \centering
    % \vspace{-2 mm}
    % \renewcommand{\arraystretch}{1.3} %调行距
    \setlength\tabcolsep{2pt} %调列距
    \begin{tabular}{c l r r r r}
        \toprule
        Dataset  & Methods & PSNR $\uparrow$  & SSIM $\uparrow$& SAM $\downarrow$ & ERGAS $\downarrow$\\
        \hline
        \multirow{2}{*}{\makecell[c]{CAVE\\32$\times$}} & Fixed    & 43.17  & 0.9927 & 4.99 & 0.203 \\
                                & Progressive       & \textbf{44.33} & \textbf{0.9951} & \textbf{3.71} & \textbf{0.179}\\
        \hline
        \multirow{2}{*}{\makecell[c]{PaviaU\\4$\times$}} & Fixed    & 45.06  & 0.9970  & 1.63 & 1.173 \\
                                & Progressive       & \textbf{46.47} & \textbf{0.9977} & \textbf{1.45} & \textbf{1.053}\\
        \hline
        \multirow{2}{*}{\makecell[c]{Chikusei\\4$\times$}} & Fixed   & 55.92 & 0.9999 & 0.50  & 0.453\\
                                    & Progressive     & \textbf{57.34} & \textbf{0.9999} & \textbf{0.43}  & \textbf{0.324}\\
        \bottomrule
    \end{tabular}
    \caption{Ablation study on progressive learning.}
    \vspace{-3mm}
    \label{Tab:Ablation-study-progressive-learning}
    % \vspace{-2.1mm}
\end{table}

Progressive learning helps CDFormer to capture long-range dependencies of spatio-spectral information in HR-HSIs.
To illustrate the effect of progressive learning, we train the CDFormer with fixed patches ($128^2$ for CAVE and Chikusei; $64^2$ for PaviaU) with the results shown under the heading of "Fixed" in Table \ref{Tab:Ablation-study-progressive-learning}.
As can be seen, progressive learning (from $128^2$ to $512^2$ for CAVE and Chikusei; from $64^2$ to $128^2$ for PaviaU) provides better results than training with fixed patches.

\vspace{-0.622mm}
\subsection{Generalization Analysis on Real Dataset}
\vspace{-0.222mm}

% In order to analyze the generalization ability, we test the performance of all competitors on the real HypSen dataset \cite{yang2018hyperspectral}.
% Due to the lack of an ideal HR-HSI to train deep neural networks, we utilize the networks trained on the PaviaU dataset to merge observed LR-HSI and the HR-MSI.
% In addition, the interpolation method is used to solve the problem of the inconsistent number of bands between datasets.
% The fusion results of all compared methods are visualized in Figure \ref{Fig:Visualized-results-HypSen}.
% It can be seen from Figure \ref{Fig:Visualized-results-HypSen} that our method generates rich details and obtain satisfactory image quality.

To examine the generalization ability of the implementations following the proposed approach, we test the performance of all competitors on the real-world HypSen dataset \cite{yang2018hyperspectral}.
Due to the lack of an ideal HR-HSI to train deep neural networks, we utilize the networks trained on the PaviaU dataset to merge observed LR-HSI and the corresponding HR-MSI.
In addition, interpolation is applied to addressing the problem of an inconsistent number of bands between datasets.
% The fusion results of all compared methods are visualized in Figure \ref{Fig:Visualized-results-HypSen}, from which it can be seen that our method generates rich details, attaining satisfactory image quality.
The fusion results of all compared methods are visualized in Figure \ref{Fig:Visualized-results-HypSen}, from which it can be seen that our method generates rich details, attaining satisfactory quality.

\vspace{-0.811mm}
\section{Conclusion}

% In this paper, we present HSR-Diff that initializes the HR-HSI with pure Gaussian noise and iteratively refines it with the condition of the LR-HSI and the HR-MSI.
% At each step, the noise is removed with CDFormer which utilizes the hierarchical representations of HR-MSI and LR-HSI rather than the original images.
% In addition, we employ a progressive learning strategy to exploit the global information of full-resolution images.
% Extensive experiments are conducted on four public datasets to validate the comparable or superior performance of the proposed method when compared with other state-of-the-art ones.
% For future work, we will try to solve the low image-generation efficiency of HSR-Diff.

In this paper, we have presented the novel HSR-Diff approach that initializes an HR-HSI with pure Gaussian noise and then, iteratively refines it subject to the condition of the LR-HSIs and HR-MSIs of interest.
At each step, the noise is removed with CDFormer which exploits the hierarchical representations of HR-MSIs and LR-HSIs rather than the original images.
% We employ a progressive learning strategy to exploit the global information of full-resolution HSIs, with CDFormer being trained on small image patches in the early epochs and on the global images in the later epochs.
In addition, we employ a progressive learning strategy to maximize the use of the global information of full-resolution images, where CDFormer is trained on small patches in the early epochs with high efficiency while on the global images in the later epochs to obtain the global statistics.
Systematic experimental investigations have been conducted, on four public datasets to validate the superior performance of the proposed approach, in comparison with state-of-the-art methods.
For future work, we will try to resolve the challenging issue of the relatively low image-generation efficiency of HSR-Diff.

\newpage

{\small
\bibliographystyle{ieee_fullname}
\bibliography{egbib}
}

\end{document}